%% file: tmlr.tex
\documentclass[10pt]{article}
\usepackage[preprint]{tmlr}

\input{math_commands.tex}

\usepackage{hyperref}
\usepackage{url}
\usepackage{amsmath,amssymb,amsthm}
\usepackage{booktabs}
\usepackage{longtable}
\usepackage{makecell}
\usepackage{graphicx}
\usepackage{xcolor}

\newtheorem{proposition}{Proposition}
\newtheorem{corollary}[proposition]{Corollary}
\theoremstyle{remark}

\newcommand{\FA}{\mathrm{FA}}                 

\title{Decomposing and Detecting Shared Error Across Models}

\author{Matt Gorbett$^{1}$ \qquad Sina Shaham$^{1}$ \qquad Suman Jana$^{2}$ \\
      {\normalfont $^{1}$Capital One \qquad $^{2}$Columbia University}}

\begin{document}
\maketitle
\thispagestyle{fancy}\lhead{Preprint}

\input{sections/abstract}

\input{sections/intro}

\input{sections/related}

\input{sections/theory}
\input{sections/method}

\input{sections/results}

\input{sections/discussion}

\input{sections/conclusion}

\bibliographystyle{tmlr}
\bibliography{tmlr}

\appendix
\input{sections/appendix_floor}

\input{sections/appendix_datasets}
\input{sections/appendix_ccs}
\input{sections/appendix_gate}
\input{sections/appendix_sweeps}

\input{sections/appendix_pooling}
\input{sections/appendix_deployment}
\input{sections/appendix_pairs}

\end{document}

%% file: math_commands.tex
\usepackage{amsmath,amsfonts,bm}

\def\eqref#1{equation~\ref{#1}}

\def\1{\bm{1}}

\DeclareMathAlphabet{\mathsfit}{\encodingdefault}{\sfdefault}{m}{sl}
\SetMathAlphabet{\mathsfit}{bold}{\encodingdefault}{\sfdefault}{bx}{n}



%% file: sections/abstract.tex
\begin{abstract}
AI oversight frequently relies on one language model auditing another, 
treating model agreement as a signal that output is correct. 
However, agreement 
conflates joint correctness with shared error, leaving identical mistakes 
unflagged. To isolate this blind spot, we analyze its cleanest formulation: 
two models independently evaluating the same fixed claims. We examine how much of this shared error can be detected without human labels, 
and evaluate whether inspecting internal representations rather than raw outputs mitigates or exacerbates the issue. 
We first formalize false agreement, defined as the situation where both models accept the same false claim,
 and show that it is determined exactly by each model's error rate and the correlation between their errors. 
 To detect these unflagged failures without ground truth, we introduce a label-free consistency floor that 
 uses complementary claim pairs (such as ``$x > y$'' versus ``$x < y$'') to establish a guaranteed lower bound on shared error. 
 Evaluating 15 open-weight models (1.7B--405B) across five datasets, we test both text outputs and activation-based linear probes. 
 We show that while probes reduce false agreement from 13\% to 9\%, this improvement stems entirely from higher accuracy, 
 and not error independence; in fact, probe errors remain just as correlated as raw text outputs.
We use the consistency floor to formalize a single risk of false agreement, the share of jointly accepted claims that are false: at matched positive rates, up to $37\%$ of what two models' outputs agree on is false, and the floor certifies at least $11\%$ without labels.
Next, we test the common assumption that cross-family pairings prevent shared errors. Our findings show this benefit is largely illusory: while cross-family pairing reduces error correlation from $0.36$ to $0.26$, its impact on false agreement is negligible and inconsistent ($9.8\%$ vs. $8.9\%$ with mass-mean probes; $8.4\%$ vs. $8.9\%$ with logistic models).
 Finally, we pose detecting false agreement as a task, evaluating supervised and label-free baselines that flag shared errors 
 among claims two models accept. 
 In-domain, a supervised detector achieves an AUROC of $0.82$ with mass-mean probes, while a score needing no labels beyond those that fit the probes, based
  on probe confidence, reaches $0.77$ with logistic probes. Performance drops to between $0.64$ and $0.66$ on unseen datasets, 
  suggesting that shared error leaves a detectable but somewhat domain-specific signature.

\end{abstract}

%% file: sections/intro.tex
\section{Introduction}
\label{sec:intro}

One language model checking another model's output is now a default instrument
of AI oversight. LLM judges grade benchmarks, filter pretraining corpora, supply
reward signals, and watch deployed behaviour. The appeal is obvious: models are
cheap, fast, and available in quantity, while human annotators are not
\citep{zheng2023judge,lambert2025rewardbench,tan2025judgebench}. Other
oversight schemes, including debate \citep{irving2018debate}, sandwiching
\citep{bowman2022measuring}, and weak-to-strong training
\citep{burns2023weak}, also rely on models checking or supervising other
models. All depend on the check remaining informative when the model being
checked is wrong.

Every such approach rests on an assumption that is almost never tested: that the checker 
and the model being checked fail differently. 
When the two agree, nothing is flagged, and agreement cannot distinguish joint correctness
from shared error. Prior work finds that model errors are substantially correlated at the output level, 
 with correlation increasing as models become more capable \citep{goel2025alike,kim2025correlated}.
 Whether the same holds for their internal representations is untested.
Additionally, measuring shared error typically requires ground-truth labels, which are absent 
in the deployment settings where oversight is most critical. In this work, we investigate 
how much shared error can be certified without ground-truth labels, and whether reading a 
model's internal representations (rather than its text outputs) makes shared error easier or harder to detect. We study the simplest form of this setting, in which two models judge the same fixed claims rather than one model judging another's generations, where errors may correlate more.

\begin{figure}[t]
\centering
\includegraphics[width=\textwidth]{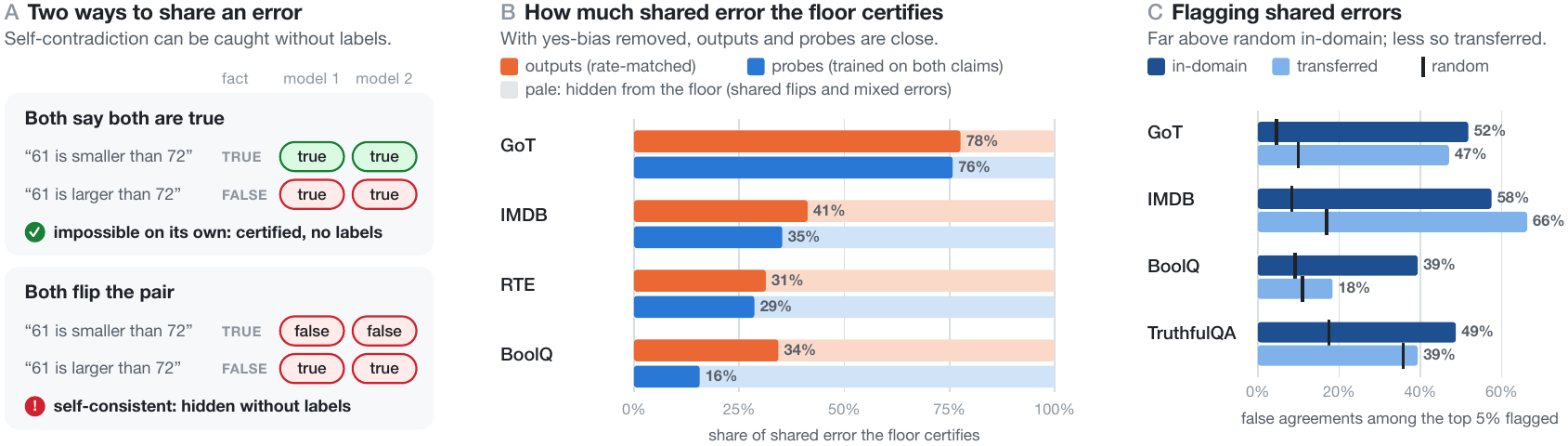}
\caption{\textbf{Shared errors that a consistency check can and cannot see.}
\textbf{(A)} A claim and its complement; exactly one is true. Green is a correct
verdict, red a wrong one. Accepting both is a self-contradiction, certified
without labels; a shared flip is equally wrong but looks like agreement on the
truth. \textbf{(B)} Of the false claims both models accept, the share the floor
certifies, i.e.\ on which both also accept the true complement ($\tau=0.5$, pooled
over model pairs); the pale rest is hidden from it. Outputs are thresholded to
their probe's positive rate rather than at $p(\textsc{yes})=0.5$, removing their
yes-bias; probes pool mass-mean and logistic, trained on both claims of each complement pair
(Table~\ref{tab:reader-design}). \textbf{(C)} Of the detector's
top-ranked $5\%$ of agreed claims, the share that are false agreements, fitted
in-domain and transferred from the other datasets; the tick is what random
review would find. Figure uses mass-mean probes. }
\label{fig:teaser}
\end{figure}

\paragraph{False agreement and a label-free floor.}
We define \emph{false agreement} as the event where two overseers both accept a false claim. 
The prevalence of false agreement depends on two factors: the error rate of each overseer
 and the extent to which their mistakes overlap.
Section~\ref{sec:theory} formalizes this relationship via classical Fr\'echet bounds, which state that false agreement cannot exceed the error rate of the superior overseer, nor can it fall below the product of their error rates unless their errors are anti-correlated. However, computing these theoretical bounds requires labels, and in practice, the Fr\'echet lower bound evaluates to zero across every probe pair we evaluate. To overcome this limitation, we introduce the \emph{consistency floor}, a lower bound on false agreement, a simple extension of single-model consistency checks \citep{fluri2024consistency}, that requires no ground-truth labels and can be measured directly on unannotated deployment data. By pairing each claim with a logical complement such that exactly one is true, any overseer that accepts both is self-contradictory and guaranteed to be wrong on at least one. When two overseers both accept both claims of a complement pair, they share a certified error (Figure~\ref{fig:teaser}). The rate of such double-acceptances establishes a non-trivial, label-free lower bound on false agreement over the target distribution.

\paragraph{Reading outputs or activations.}
Overseers that examine only generated text miss information that remains unexpressed in model activations. Linear probes recover truth-related directions in activations \citep{burns2023ccs,marks2024geometry} and can outperform text confidence during model failure \citep{liu2023dissonance}. Crucially, correlated text outputs do not imply correlated representations, as models can reach identical outputs through distinct internal paths. We compare text-output overseers directly against activation probes on the same claims. Probes yield lower absolute false agreement, but this stems from higher overall accuracy rather than error independence. Probe errors are at least as correlated as output errors and harder to detect. While text-level failures manifest as self-contradictions that the consistency floor easily catches, probes trained on both claims of each complement pair frequently flip the entire pair (accepting the false claim while rejecting the true one). This joint flip maintains internal consistency, rendering the error invisible to contradiction checks.

\paragraph{Detecting what agreement hides.}
The floor certifies that shared error exists but cannot say which agreed claims
hide the rest of it. We therefore train a \emph{false-agreement detector} on both
models' activations that ranks the claims both overseers accept by how likely
they are to be shared errors, and ask whether that signal survives a change of
distribution and whether a score needing no labels can match it.

\paragraph{Contributions.}
The study covers $498$ unique model-pair--dataset cells formed from $15$
open-weight models across five families, spanning $1.7$B to $405$B parameters,
on five datasets. It makes
four contributions.

\begin{itemize}
\itemsep0em
\item \textbf{Certifying shared error without labels.} We introduce the consistency floor, a label-free lower bound on false agreement derived from complement pairs alone. False agreement divides cleanly into visible contradictions (certified by the floor) and invisible shared flips (where both models accept the false claim and reject the true one). 

\item \textbf{Comparing outputs and internal readouts.} At matched positive rates, probes lower false agreement from $13.3\%$ to $9.1\%$ of false claims, but through accuracy rather than independence: probe errors are at least as correlated as output errors ($\rho=0.28$ against $0.25$). What the floor certifies depends on the specific estimator rather than whether one inspects text or activations: with yes-bias removed, outputs and probes certify comparable shares ($30$--$78\%$ and $5$--$77\%$), and the rest are shared flips that no contradiction check can see.
\item \textbf{Model-family diversity lowers error correlation but not the blind spot.} Pairing across families lowers error correlation consistently: $0.36$ to $0.26$ with mass-mean probes ($t=-5.0$) and $0.42$ to $0.35$ with logistic ones ($t=-4.0$). However, it does not shrink the blind spot. False agreement is $9.8\%$ within families against $8.9\%$ across them with mass-mean probes, and $8.4\%$ against $8.9\%$ with logistic ones. The decomposition says why: correlation governs about half of false agreement, so a reduction of this size is worth roughly one percentage point, against pair-to-pair variation spanning $0.5\%$ to $29\%$. Choosing an overseer by family is choosing on the weaker of the two terms.
\item \textbf{Detecting false agreement: a task and baselines.} We pose flagging shared errors among agreed claims as a benchmark task and evaluate supervised and label-free baselines. The strongest approach does not rely on joint representations. Activations from a single model perform as well as a classifier trained on both, and a standard truth probe surpasses it. Shared errors are found by better readers rather than joint structure.
\end{itemize}

%% file: sections/related.tex
\section{Related Work}
\label{sec:related}

\paragraph{Model judges and correlated output errors.}
Language models are widely used to grade outputs, supply feedback, and supervise other models \citep{zheng2023judge,irving2018debate,bowman2022measuring,burns2023weak}, a paradigm that underlies most proposals for scalable oversight \citep{amodei2016concrete}. It assumes a judge stays informative when the evaluated model is wrong, and recent work challenges that at the output level. Specifically, judges prefer their own generations \citep{panickssery2024selfpreference}, output similarity degrades weak-to-strong supervision \citep{goel2025alike}, models agree substantially on incorrect answers \citep{kim2025correlated}, and agreement does not imply correctness \citep{ding2026agree}. Ensembling more judges helps less than their number suggests. A panel of nine
frontier judges carries only about two votes' worth of independent information
\citep{kohli2026nine}, the gain from routing or voting is capped by how often
every model fails at once \citep{chen2026cofailure}, and error common to all
evaluators survives averaging and cannot be seen in their scores
\citep{afrin2026ensembles}. \citet{kim2025correlated} find output-error agreement
is higher within model families, and we test whether that pattern holds for
internal readouts.

\paragraph{Truth probes and internal readouts.}
Linear probes can isolate truth-related directions within model activations \citep{burns2023ccs,marks2024geometry} and outperform expressed text confidence when outputs are wrong \citep{liu2023dissonance}. However, probe interpretation requires care. Unsupervised probing can recover salient binary features unrelated to truth \citep{farquhar2024challenges}, and linear truth directions do not always transfer across tasks or logical forms \citep{bao2025probing}. We address this by validating probes against output confidence and focusing our main results on supervised linear classifiers. While prior work identifies internal signals of user sycophancy \citep{pandey2026sycophancy}, our work focuses on a distinct unit of analysis, evaluating the joint readouts of two separate models and the conditions under which both fail simultaneously.

\paragraph{Error correlation and ensemble diversity.}
The dependence of ensemble performance on error covariance is well-established \citep{brown2005diversity}, where false agreement is the \emph{double-fault} measure \citep{giacinto2001,kuncheva2003} applied to internal readouts rather than classifier outputs. Similarly, Fr\'echet bounds define how marginal error rates constrain potential event overlap \citep{frechet1951tableaux}. Rather than introducing a new similarity metric, we leverage these classical tools to decompose probe oversight failure into individual error rates versus correlated error. This theoretical decomposition reveals the bounds on false agreement imposed by probe accuracy, explaining why lower error correlation does not necessarily reduce absolute error overlap.

\paragraph{Consistency checks and label-free error estimates.}
Logical consistency provides evidence of error without ground truth, as a model asserting both a claim and its negation is necessarily wrong on one \citep{fluri2024consistency}. Prior work notes that language models frequently exhibit inconsistency under paraphrase \citep{elazar2021consistency} and negation \citep{kassner2020negated}. Consistency constraints also guide truth probing, as seen when CCS enforces complementary scores for negated claims \citep{burns2023ccs}, though supervised probes often fail on negated statements \citep{levinstein2024still,marks2024geometry}. 

A separate line of work estimates classifier error rates from unlabeled agreement data \citep{dawid1979maximum,platanios2014estimating,parisi2014ranking,baek2022agreement}, extending to logical output constraints \citep{platanios2017estimating} and dependent classifiers \citep{jaffe2016unsupervised}. These estimators assume specific error co-occurrence models (typically conditionally independent errors given the label), which breaks down precisely when models share errors. We invert this relationship: instead of estimating error rates from agreement, we count shared contradictions. This yields a label-free lower bound on shared error without assuming error independence (Section~\ref{sec:theory:floor}). We also delineate where this check fails, showing that internally consistent shared errors leave no contradiction, causing agreement-based upper bounds to break down under correlated error.

\paragraph{Detecting disagreement and shared error.}
Prior work uses cross-model surprise to detect confident errors beyond within-model uncertainty baselines \citep{gorbett2026disagreement}, while probe-based uncertainty methods estimate confidence from single-judge activations \citep{radharapu2026calibrating,bao2025probing}. We study the harder problem of two agreeing readers. Specifically, we test whether joint activations reveal false agreement, how well that signal transfers across domains, and whether a label-free score based on the smaller probe margin can match supervised detection. While \citet{afrin2026ensembles} show that common-mode evaluator error cannot be identified from output scores alone, we demonstrate when and how internal activations make it detectable.

%% file: sections/theory.tex
\section{What Agreement Can Catch}
\label{sec:theory}

Agreement between two readers is informative only when they do not fail simultaneously. This section formalizes that condition, demonstrates that standard statistical bounds on shared error require ground-truth labels, and introduces a lower bound that does not.

Consider the evaluation set of claims whose ground truth is \textsc{false}. Let $E_i = 1$ when reader $i$ (either a model's text output or a linear probe on its activations) incorrectly classifies such a claim as \textsc{true}, and let $p_i = \mathbb{P}(E_i = 1)$ denote its false-positive rate. The false-agreement rate is defined as:
\begin{equation}
  \text{FA} = \mathbb{P}(E_1 = 1, E_2 = 1).
  \label{eq:fa}
\end{equation}
This quantity represents the blind spot of disagreement-based oversight: because both readers produce identical errors, neither raises an objection.

\subsection{False agreement and its bounds}
\label{sec:theory:identity}

Let $\rho$ denote the Pearson correlation coefficient between the binary error indicators $E_1$ and $E_2$. Expanding the covariance $\text{Cov}(E_1, E_2) = \mathbb{E}[E_1 E_2] - \mathbb{E}[E_1]\mathbb{E}[E_2]$ yields:
\begin{equation}
  \text{FA} = p_1 p_2 + \rho \sqrt{p_1 (1 - p_1)} \sqrt{p_2 (1 - p_2)}.
  \label{eq:identity}
\end{equation}
The first term represents the expected overlap under error independence, while the second term captures the additional overlap from correlated errors. While lower individual error rates shrink the independence term and greater diversity reduces the correlation term, decreasing one does not guarantee a smaller sum. A more diverse overseer may trade lower correlation for a higher overall error rate. Moreover, the correlation term is often small relative to the independent term, meaning a reduction in correlation may have a minimal impact on total false agreement.

The marginal error rates also strictly constrain error overlap. The classical Fr\'echet bounds require:
\begin{equation}
  \max(0, p_1 + p_2 - 1) \leq \text{FA} \leq \min(p_1, p_2).
  \label{eq:frechet}
\end{equation}
The upper bound corresponds to the error rate of the more accurate reader, reached when the other shares all of its errors. In our evaluations, the Fr\'echet lower bound is zero for every probe pair, as it must be whenever $p_1+p_2<1$. For output pairs, it is positive in just over half of the cases, driven by a systematic \textsc{yes}-bias that pushes $p_1 + p_2 > 1$. Independent errors lie within this theoretical interval. For instance, with marginal error rates of $10\%$ and $5\%$, independent errors yield $0.5\%$ false agreement against a theoretical maximum of $5\%$, presenting a tenfold variation at fixed accuracy. 

These bounds also determine the coverage $C$ of a disagreement router (the fraction of the more accurate reader's errors that the second reader exposes by disagreeing):
\begin{equation}
  C = 1 - \frac{\text{FA}}{\min(p_1, p_2)},
  \label{eq:coverage}
\end{equation}
which degrades to zero at the Fr\'echet upper bound.

To isolate error overlap relative to its mathematical maximum, we report normalized correlation:
\begin{equation}
  \tilde{\rho} = \frac{\rho}{\rho_{\max}},
  \qquad
  \rho_{\max} = \frac{\min(p_1, p_2) - p_1 p_2}{\sqrt{p_1 (1 - p_1) p_2 (1 - p_2)}},
  \label{eq:rhonorm}
\end{equation}
which measures the fraction of maximum attainable error overlap permitted by the marginals (with negative correlations normalized by their feasible minimum). These relations follow from binary covariance and Fr\'echet inequalities \citep{frechet1951tableaux}. They illustrate what labeled data can certify, establishing a baseline for what can be achieved without labels in Section~\ref{sec:theory:floor}.

\begin{figure}[t]
\centering
\includegraphics[width=\textwidth]{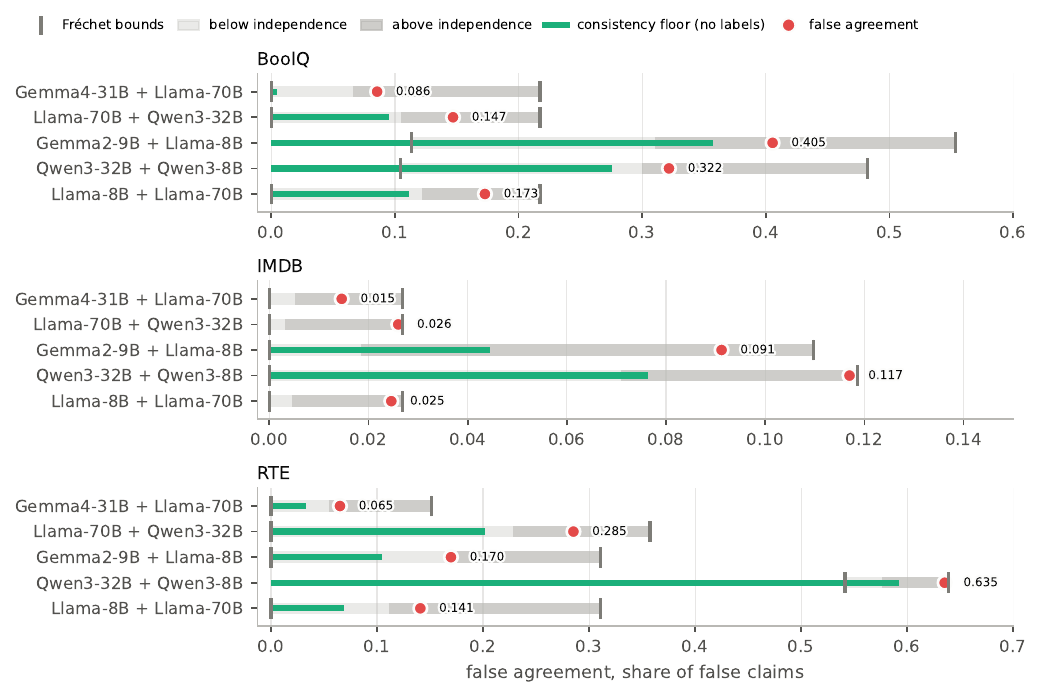}
\caption{\textbf{Five model pairs against the interval permitted by their error rates, alongside the certified consistency floor.} Model outputs across three datasets for five fixed model pairs. Each bar represents a pair's Fr\'echet interval (Equation~\ref{eq:frechet}), with gray ticks marking $\max(0, p_1 + p_2 - 1)$ and $\min(p_1, p_2)$. Shading indicates values below independence ($p_1 p_2$, pale) and above independence (darker, indicating positive error correlation). The green segment extending from zero represents the consistency floor (Section~\ref{sec:theory:floor}) certified without labels; red dots denote measured false agreement. All values are computed on complement pairs ($\tau=0.5$). Outputs are read at $p(\textsc{yes})>0.5$, which is yes-biased and inflates the floor's share; Table~\ref{tab:reader-design} removes this bias. Figure~\ref{fig:bound-probe} presents corresponding results for mass-mean probes.}
\label{fig:bound}
\end{figure}

\subsection{A bound that needs no labels}
\label{sec:theory:floor}

Evaluating $p_1$, $p_2$, and $\text{FA}$ requires ground-truth labels because all three are defined over false claims. Furthermore, because the Fr\'echet lower bound is zero across all probe pairs, label-based theoretical lower bounds offer little practical value. We address this with a lower bound that needs no labels, and show it is the tightest one available.

We pair each false claim $\bar{s}$ with a logical complement $s$ such that exactly one statement is true, and evaluate each reader on both statements independently. When presented with a complement pair, a reader that accepts the false statement $\bar{s}$ commits one of two mutually exclusive error types: it accepts both statements ($A_i$), or it accepts the false statement while rejecting the true statement ($F_i$, a \emph{flip}). Because these events are disjoint, the error decomposes as $E_i = A_i \cup F_i$, yielding:
\begin{equation}
  \text{FA}
  = \underbrace{\mathbb{P}(A_1, A_2)}_{\text{CF}}
  + \mathbb{P}(A_1, F_2) + \mathbb{P}(F_1, A_2)
  + \mathbb{P}(F_1, F_2).
  \label{eq:floor}
\end{equation}

The first term, $\mathbb{P}(A_1, A_2)$, can be computed entirely without ground-truth labels. A reader accepting both claims of a complement pair is self-contradictory and guaranteed to be wrong on at least one. Consequently, instances where both readers accept both statements represent certified shared errors. We define their rate as the \emph{consistency floor}:
\begin{equation}
  \text{CF} = \mathbb{P}(A_1, A_2) \leq \text{FA}.
\end{equation}
Because each complement pair contains exactly one false claim, the total number of pairs equals the number of false claims, allowing $\text{CF}$ to share the exact denominator as $\text{FA}$.

For example, consider the complement pair \emph{``Sixty-one is smaller than seventy-two''} and \emph{``Sixty-one is larger than seventy-two''} from Geometry of Truth. A reader accepting both is self-contradictory, a fact verifiable without evaluating arithmetic. Conversely, a reader accepting \emph{larger} while rejecting \emph{smaller} commits a flip: its answer is incorrect but structurally indistinguishable from a confident, correct verdict. If both readers evaluate $100$ such pairs and accept the false statement on $10$ pairs, with $6$ of those pairs also involving acceptance of the true statement, then $\text{FA} = 10\%$ and $\text{CF} = 6\%$. The 6 double-acceptances are certified without labels, while the 4 flip-based errors remain indistinguishable from genuine agreement on truth.

The remaining three terms in Equation~\ref{eq:floor} cannot be measured without labels. A flip is internally consistent (accepting the false statement while rejecting the true one), making it visually identical to a correct verdict from model outputs alone. Thus, $\text{CF}$ equals $\text{FA}$ only when no shared errors involve flips. The gap $\text{FA} - \text{CF}$ measures the fraction of shared errors invisible to contradiction checks, and it is widest for the readers one would most want to deploy: a reader trained to answer complements consistently rarely contradicts itself, so its errors are almost all flips and its floor approaches zero whatever its false agreement (Table~\ref{tab:reader-design}). The floor therefore audits a specific failure, self-contradiction, rather than bounding shared error in general. Unlike the Fr\'echet lower bound, $\text{CF}$ can be positive even when $p_1 + p_2 < 1$, certifying shared errors in regimes where marginal bounds fail to provide guarantee.

The following observations are elementary, but they fix what can be certified. Without labels, CF is the most that can be said from below, and the only bound from above is loose. Write $a_i, \bar a_i \in \{0,1\}$ for whether reader $i$ accepts the two claims of a pair; these are observed, and which claim is true is not.

\begin{proposition}[Sharp label-free bounds]
\label{prop:sharp}
For every assignment of truth to the claims of the $N$ pairs, the false-agreement rate lies in $[\mathrm{CF}, \mathrm{U}]$, where $\mathrm{CF} = \frac1N\#\{a_1=\bar a_1=a_2=\bar a_2=1\}$ and $\mathrm{U} = \frac1N\#\{(a_1 \wedge a_2) \vee (\bar a_1 \wedge \bar a_2)\}$, the rate at which both readers accept a common claim. Both endpoints are attainable, so the bounds are sharp.
\end{proposition}

\begin{proof}
Assignments are free per pair, so it suffices to consider one pair. A false agreement needs both readers to accept the false claim. If both accept both claims, this holds whichever claim is false. If exactly one claim is accepted by both, it holds when that claim is false and fails when it is true. If no claim is accepted by both, it fails under either assignment. Taking the favourable or unfavourable assignment on every pair attains CF and U.
\end{proof}

\begin{corollary}
\label{cor:flip}
The shared-flip rate $\mathbb{P}(F_1, F_2)$ is not identified from unlabelled verdicts: relabelling a pair turns agreement on the true claim into a shared flip without changing any observation.
\end{corollary}

Additional unlabelled statistics cannot raise the floor without assumptions about which claim is true, such as the conditional independence that agreement-based estimators assume.

\paragraph{Risk of false agreement.}
Oversight acts on what two readers agree on, so we summarise shared error as a single risk: of the pairs on which both readers accept a common claim, the share on which that claim is false,
\begin{equation}
  \mathrm{RFA} = \frac{\mathrm{FA}}{\mathrm{U}}, \qquad \frac{\mathrm{CF}}{\mathrm{U}} \leq \mathrm{RFA} \leq 1.
  \label{eq:rfa}
\end{equation}
Both $\mathrm{CF}$ and $\mathrm{U}$ are observed, so the lower bound needs no labels; by Corollary~\ref{cor:flip}, no upper bound below $1$ can be certified.

This formulation extends contradiction-based validation from single models to multi-model oversight. While single-model contradictions provide label-free evidence of individual error \citep{fluri2024consistency}, $\text{CF}$ isolates joint contradictions to certify shared errors without making independence assumptions. 

The Dawid--Skene baseline illustrates the limits of agreement-based error estimation under shared errors; we describe it in Appendix~\ref{app:floor:ds}.

%% file: sections/method.tex
\section{Method}
\label{sec:method}

We compare two readers for each model, its generated text output and a linear probe on its activations, across three core evaluations: false agreement, the consistency floor, and a false-agreement detector that scores claims accepted by both readers. All fitted components are evaluated strictly on held-out test data.

\subsection{Readers}
\label{sec:method:probe}

For any given model and claim, we construct a contrast pair by appending \texttt{Is this claim true? Yes} and \texttt{No} to the claim. Because these prompts differ by a single token, variations in the resulting hidden states capture properties of the claim itself rather than artifacts of surface phrasing. We extract the final-layer hidden state at the last token of each half, the \texttt{Yes} or \texttt{No} token itself; the two halves share every earlier state, so their difference is carried by that token. Appendix~\ref{app:pooling} justifies our focus on the final layer over mean-pooling across the prompt.

The \emph{output reader} evaluates the model's native $P(\text{Yes})$ probability directly without fitting parameters. It is a zero-shot readout, not a templated or chain-of-thought judge. The \emph{probe readers} apply linear classifiers to identical activations, differing only in how the classification direction is determined:
\begin{itemize}
\itemsep0em
\item \textbf{Mass-mean}: The difference between class means, which \citet{marks2024geometry} find has stronger causal effects under intervention than other probe directions.
\item \textbf{Logistic regression}: Standard linear probing fit on the same features.
\end{itemize}
Each probe's projection is divided by its spread on the training set and passed through a sigmoid, so a score of $0.5$ is the probe's fitted decision boundary: the midpoint between class means for mass-mean, the logistic boundary for logistic. An unsupervised baseline, Contrast-Consistent Search (CCS) \citep{burns2023ccs}, is analyzed in Appendix~\ref{app:ccs}. Probes are read at the final layer; Appendix~\ref{sec:results:depth} reports the same measurements at $0.5$ and $0.75$ of model depth.

To ensure probes capture genuine latent structure rather than merely re-encoding expressed text confidence, we benchmark probe scores against model output confidence on high-confidence items (Appendix~\ref{app:gate}).

\subsection{The consistency floor}
\label{sec:method:floor}

\paragraph{Complement pairs and extraction.}
We implement the consistency floor by evaluating models on complementary pairs $(s, \bar{s})$ where exactly one statement is true. We call $s$ and $\bar{s}$ the two \emph{claims} of a complement pair, keeping \emph{halves} for the \texttt{Yes}/\texttt{No} prompts of a contrast pair. Across benchmarks, complements are derived naturally: dataset-native negations for Geometry of Truth, matched exactly on the entities each statement names (the companies topic has none and is excluded), \texttt{Yes}/\texttt{No} answers for BoolQ, entailment wrappers for RTE, and polarity classes for IMDB. All four constructions are mechanical and need no labels. TruthfulQA is excluded: pairing a correct with an incorrect answer requires its labels, and its answers are not logical complements. Prompts exceeding remote model windows ($512$ tokens) are excluded to preserve token symmetry. Complements are of three kinds. Geometry of Truth supplies \emph{logical complements}, a claim and its negation. BoolQ and RTE give \emph{label-complementary} pairs, where the benchmark's own labelling makes exactly one member true. IMDB gives a \emph{polarity contrast}, complementary under the dataset's binary labelling but not a negation. The floor's guarantee needs only that exactly one member is true, which holds in all three, but the three are not equally strong evidence that a reader understood the claim.

\paragraph{Counting and fitting protocol.}
A reader accepts a claim when its score exceeds a decision threshold $\tau = 0.5$. The consistency floor ($\text{CF}$) measures the fraction of $N$ pairs where both models accept both claims, accompanied by Clopper--Pearson 95\% confidence intervals \citep{clopper1934use}. Ground-truth labels are applied afterward solely to compute $\text{FA}$ and decompose shared errors via Equation~\ref{eq:floor}. Probes are trained on both claims of each training pair, with 5-fold cross-fitting over pairs to ensure zero data leakage. Two controls separate reader design from reader type: probes trained on the original claim only, and outputs thresholded to their probe's positive rate.

\paragraph{Label-free ceiling baseline.}
We compare the floor with a Dawid--Skene agreement-based ceiling estimate; its construction and limitations are described in Appendix~\ref{app:floor:ds}.

\subsection{The false-agreement detector}
\label{sec:method:direction}

As a supervised baseline, we train a linear classifier, the \emph{false-agreement detector}, to separate genuine truths from shared errors among claims accepted by both probes. We also score the same claims with each model's logistic truth probe, trained on truth labels rather than false-agreement labels. For each model, we extract mean and difference activation vectors from the \texttt{Yes} and \texttt{No} halves of the contrast pair. Concatenating these four vectors (mean and difference per model) forms a joint feature vector. Each feature is standardized, and the detector is an L2-regularized logistic regression with $C = 1$, not tuned. 

Retaining the mean activation preserves shared properties such as text fluency and familiarity, which may explain why models co-accept false claims. The difference vector isolates sensitivity to the polarity suffix. Performance is evaluated against a permutation control (shuffled labels).

\paragraph{Label-free ranking alternatives.}
\label{sec:method:unsup}
Because oversight environments lack ground-truth labels, we evaluate whether label-free scores can rank agreed claims effectively. We examine two label-free heuristics:
\begin{itemize}
\itemsep0em
\item \textbf{Margin}: $-\min(|b_1 - 0.5|, |b_2 - 0.5|)$, capturing proximity to the decision boundary.
\item \textbf{Anomaly}: The Mahalanobis distance of joint activations from the pool center.
\end{itemize}
Appendix~\ref{app:labelfree} provides full definitions. Here and for the floor, label-free means the check itself uses no labels; the probes it reads are trained on labeled data.

\subsection{Protocol}
\label{sec:method:protocol}

Data splits are group-aware to prevent source-question leakage across train and test sets. To test generalization under distribution shift, we employ a leave-one-dataset-out protocol where probes and detectors are trained on four datasets and evaluated entirely on the held-out fifth. To account for shifts in probe score distributions on unseen datasets, we dynamically recalibrate cutoffs to match each model's training positive rate without peeking at test labels.

\paragraph{Models, data, and compute.}
Our evaluation spans 15 open-weight models across five families (Gemma, Llama, OLMo, Qwen, GPT-J) ranging from $1.7\text{B}$ to $405\text{B}$ parameters, yielding $498$ model-pair--dataset cells across five benchmarks: Geometry of Truth \citep{marks2024geometry}, TruthfulQA \citep{lin2022truthfulqa}, BoolQ \citep{clark2019boolq}, RTE \citep{wang2018glue}, and IMDB \citep{maas2011imdb}. Activations are cached in \texttt{fp16}, making all downstream analyses deterministic and accelerator-independent. For hybrid thinking models (6 of 15), reasoning tokens are disabled to probe pre-reasoning beliefs.

\paragraph{Shared training data.}
Both probes in a pair are fitted on the same items, so two linear fits could
absorb the same sample noise and inflate their error correlation. Refitting each
on a disjoint, group-aware half of the training split lowers $\rho$ by $0.043$
for logistic probes ($t=-10.9$ over $611$ cells) and leaves mass-mean unchanged
($+0.008$, $p=0.05$), with false agreement unmoved. Halving the fitting data does
raise the better probe's error rate slightly ($0.186$ to $0.210$ for logistic),
so this is an upper bound on what the shared split contributes: part of the
logistic correlation, and none of the reported false agreement.

\paragraph{Which grid produces which table.}
The consistency floor is computed on the complement-pair datasets with probes cross-fitted within a pair; every other table uses the five-dataset grid and the seed-0 split, so the two sets of false-agreement rates are not interchangeable.

\paragraph{Scope limitation.}
The floor also needs complements to exist: ours are constructed from each dataset's own structure, whereas a deployment would generate them, and a malformed pair in which both claims are true or both false breaks the guarantee rather than merely weakening it. Our analysis measures agreement between model readers rather than underlying cognitive beliefs. Shared false agreements may reflect genuine shared misconceptions or common vulnerabilities to surface-level text familiarity.

%% file: sections/results.tex
\section{Results}
\label{sec:results}

\subsection{Shared error certified without labels}
\label{sec:results:floor}

\input{tables/consistency_floor.tex}

We begin with what can be measured without labels. Table~\ref{tab:consistency-floor} sets the consistency floor (Section~\ref{sec:theory:floor}) against every other bound. Where the Fr\'echet lower bound is zero, which is every probe cell, the floor is not: its 95\% interval excludes zero in most model pairs. In some cells, the floor alone exceeds the independence reference $p_1 p_2$ (most often IMDB with logistic probes), meaning correlated error is certified by contradictions without any joint labels.

The floor recovers most of the output-level blind spot ($71$--$98\%$ of false agreement), but much less of the probe-level blind spot ($5$--$39\%$ on RTE, BoolQ, and IMDB). Splitting each false agreement by how the two readers reached it explains the gap. At the output level, both readers accepted both claims in $71$--$98\%$ of cases and both flipped the pair in at most $5\%$. At the probe level, on RTE, BoolQ, and IMDB, both flipped the pair in $23$--$73\%$, and a shared flip is a consistent answer that no contradiction can expose. Part of this difference may stem from the reader rather than the model: the probes are fitted on both claims of each pair, which pushes them toward consistent answers, while the outputs carry a bias toward \textsc{yes}. Controlling for both narrows the gap: outputs thresholded to the probes' positive rate recover $30$--$78\%$, and probes fitted on the original claim only recover $20$--$98\%$ (Table~\ref{tab:reader-design}).

The Dawid--Skene estimate falls below measured false agreement in most IMDB cells for every reader (Appendix~\ref{app:floor:ds}). The sharp label-free ceiling of Proposition~\ref{prop:sharp}, the rate at which both readers accept a common claim, is valid but loose: its median is $71$--$94\%$ for outputs and $56$--$99\%$ for probes, against false agreement of $2$--$33\%$ and $0$--$15\%$. Expressed as risk (Equation~\ref{eq:rfa}, median over model pairs), with outputs thresholded to their probe's positive rate to remove yes-bias (Table~\ref{tab:reader-design}), $8$--$37\%$ of what two models' outputs jointly accept on RTE, BoolQ and IMDB is false, and at least $2$--$11\%$ is certified without labels. Probes carry less risk, $5$--$22\%$, and less of it is certifiable, $0.5$--$5\%$. On GoT the risk is below $3\%$ for every reader.

\subsection{Probes share fewer errors through accuracy, not independence}
\label{sec:results:diversity}

How much shared error the floor sees depends on reader design; whether probes share less of it is a separate question. Figures~\ref{fig:bound} (outputs) and~\ref{fig:bound-probe} (probes) place the measured rate inside the Fr\'echet interval derived in Section~\ref{sec:theory}. Across dataset averages, false agreement ranges from $\mathbf{1.4\%}$ to $\mathbf{16.8\%}$ with logistic probes and from $\mathbf{4.0\%}$ to $\mathbf{12.9\%}$ with mass-mean probes. Over all cells it is $9.1\%$ $[8.1, 10.2]$ with mass-mean and $8.8\%$ $[7.7, 10.1]$ with logistic probes ($95\%$ model-bootstrap intervals). \textbf{Every dataset average exceeds its independence reference}, and errors are positively correlated in $\mathbf{100\%}$ and $\mathbf{89\%}$ of the $498$ model-pair--dataset cells, respectively ($95\%$ model-bootstrap interval for the latter $[78, 99]\%$).

\paragraph{Do probes share fewer errors than outputs?}
Each model's single-token $p(\text{Yes})$ provides an output-level verdict on the same claims, allowing false agreement to be computed for output pairs on identical items and splits. At each model's native threshold, the output pair reaches $30.2\%$ false agreement, but it also labels claims true far more often than the probes do. A reader that outputs \textsc{true} more frequently incurs higher false agreement by default. Holding each output's positive rate equal to that of the probe it replaces, output-level false agreement is $13.3\%$ against the mass-mean probes' $9.1\%$, and $14.5\%$ against the logistic probes' $8.8\%$ (differences $4.2$ and $5.7$ points, $95\%$ model-bootstrap intervals $[2.9, 5.8]$ and $[4.4, 7.3]$). Dropping any one model family leaves the difference between $3.7$ and $4.8$ points with mass-mean probes and $5.3$ and $6.4$ with logistic probes. 

This gain is concentrated rather than uniform: on TruthfulQA the probes roughly halve false agreement ($10.2\%$ vs. $21.0\%$) and on RTE they cut it by a third ($12.9\%$ vs. $20.7\%$), whereas on Geometry of Truth the two are indistinguishable, and on BoolQ the mass-mean probes perform slightly worse. The gain is almost all accuracy. With mass-mean probes the independence term $p_1p_2$ falls from $9.3\%$ to $5.4\%$, while the correlation term moves from $4.0\%$ to $3.7\%$; with logistic probes, $10.5\%$ to $4.6\%$ against $4.0\%$ to $4.2\%$. Probe errors are at least as correlated as matched output errors: the gap in $\rho$ is $+0.03$ $[-0.03, 0.10]$ with mass-mean probes ($0.28$ against $0.25$; normalized, $\tilde\rho = 0.36$ against $0.34$) and $+0.12$ $[0.04, 0.19]$ with logistic probes. Label noise is unlikely to explain the gap, since two accurate probes would both ``err'' on mislabelled items: the gap is largest on Geometry of Truth, whose curated statements are nearly noise-free ($0.36$ against $0.28$), and reversed on TruthfulQA, the noisiest ($0.19$ against $0.24$). Note that this comparison is asymmetric, as probes are fitted on labeled data while outputs are read zero-shot, representing a deployment comparison rather than a controlled experiment. Supervising the output reader would not remove the asymmetry: $p(\textsc{yes})$ is a single number, so any monotone calibration of it only moves its threshold, which the matched-rate comparison already sets. A supervised classifier on richer output-side features would be a probe in all but name.

Two quantities from Section~\ref{sec:theory} summarize how much of the available overlap is realized. Normalized against the largest overlap its marginals permit, $\tilde{\rho}$ averages $0.48$ with logistic probes and $0.36$ with mass-mean probes, indicating that probe errors achieve roughly half their theoretical overlap. Coverage $C$, representing the share of the more accurate probe's errors exposed by disagreement, averages $0.39$ and $0.43$. Between one-third and two-fifths of the better reader's mistakes are exposed to the second reader, while the remainder are shared and silent.

\paragraph{Family diversity barely reduces shared error.}
Over all pairs, with logistic probes, false agreement is $8.4\%$ within families and $8.9\%$ across families; with mass-mean probes, it is $9.8\%$ and $8.9\%$ ($p = 0.32$ and $0.045$). Both differences are under one point, and their signs disagree; bootstrapping over models, the data exclude cross-family reductions larger than $1.8$ and $0.4$ points respectively.

\begin{figure}[t]
\centering
\includegraphics[width=\textwidth]{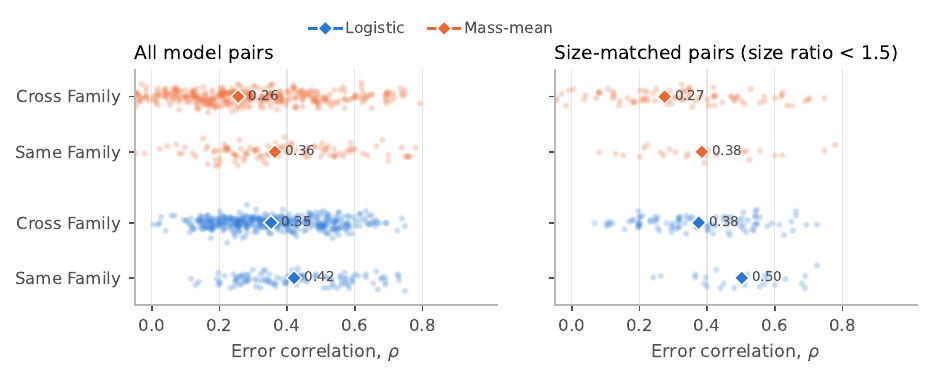}
\caption{Pair-level error correlations for within-family (same) and cross-family model pairs, shown for logistic and mass-mean probes. Dots represent model-pair--dataset cells; diamonds and labels mark group means. The left panel shows all pairs. The right panel restricts to pairs whose model sizes differ by less than a factor of $1.5$. Cross-family pairs exhibit lower mean error correlation overall, while size-matched distributions overlap substantially.}
\label{fig:diversity}
\end{figure}

Figure~\ref{fig:diversity} shows that pair-level error correlation is lower across families in the aggregate. The difference in means is $-0.068$ with logistic probes and $-0.108$ with mass-mean probes ($p = 0.004$ and $0.006$). Because the cells share models, every $p$-value in this section comes from permuting family labels across models, so all cells involving a model move together.

However, a key confound must be addressed. Two of our five families contain a single model, meaning those models appear exclusively in cross-family pairs and happen to be less accurate than average, artificially inflating the cross-family error rate. Restricting the comparison to the three families with more than one model, so every model is eligible for both pairing types, removes the accuracy gap entirely (mean error rates of $0.174$ vs. $0.167$ for logistic probes, $0.213$ vs. $0.208$ for mass-mean), while the correlation reduction survives: $0.421$ to $0.378$ ($p = 0.018$) and $0.364$ to $0.295$ ($p = 0.017$). Restricting further to size-matched pairs leaves $72$ cells, where it does not ($p = 0.075$ and $0.56$).

The correlation reduction is real but buys little, because correlation governs only part of false agreement. Within this restriction, the correlation term falls from $4.6\%$ to $4.3\%$ of false claims with logistic probes and from $4.9\%$ to $3.8\%$ with mass-mean probes, a shift of about one percentage point. Meanwhile, the independence term $p_1 p_2$ remains stable at $3.8\%$ to $4.1\%$ and $4.9\%$ to $5.1\%$. In this subset, false agreement lands at $8.4\%$ versus $8.4\%$ with logistic probes ($p = 0.92$) and $9.8\%$ versus $8.9\%$ with mass-mean probes ($p = 0.015$), a one-point gain that appears with one probe only, does not survive size matching ($p = 0.40$), and is small against pair-to-pair variation of $0.5\%$ to $29\%$.

Pair-level outcomes vary substantially within each dataset, so dataset averages do not predict how much shared error a particular pair has or how detectable it will be. The full pair-selection analysis and all $498$ cells appear in Appendix~\ref{app:pairs}.

\subsection{The false-agreement detector}
\label{sec:indomain}

The consistency floor certifies that shared error exists, but it cannot identify which agreed claims hide the remaining errors. The false-agreement detector (Section~\ref{sec:method:direction}) tests whether shared error is linearly detectable and whether that signature generalizes beyond its training distribution. Evaluating the detector inside a single dataset answers the first question. For each model pair and dataset, we fit probes on training rows and score the detector using five-fold cross-validation over remaining rows, ensuring no claim is ranked by a detector that has seen it.

\begin{figure}[t]
\centering
\includegraphics[width=\textwidth]{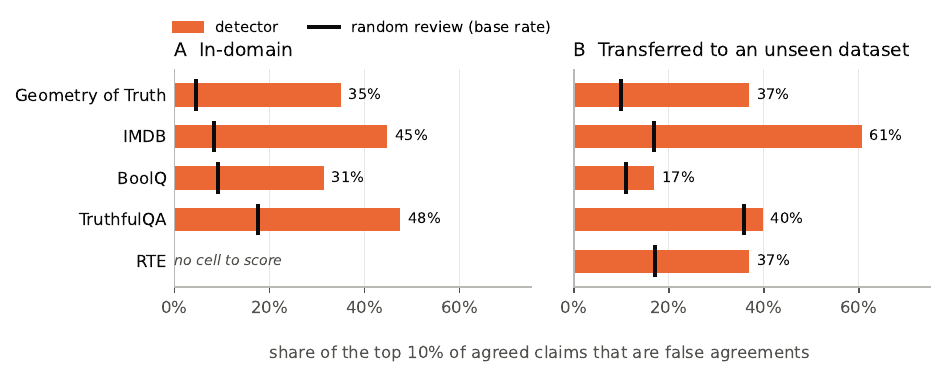}
\caption{What a $10\%$ review budget finds. Bars are the share of the detector's top-ranked $10\%$ of agreed claims (those both probes accept) that are false agreements; the tick is the base rate, what reviewing the same number of claims at random would find. Mass-mean probes, mean over cells with at least $20$ false agreements to score. \textbf{(A)} Detector fitted within the dataset; RTE has no cell to score. \textbf{(B)} Detector fitted on the other four datasets. Base rates differ between panels because transferred probes run at a recalibrated threshold, so compare lift over the tick rather than bar heights across panels.}
\label{fig:indomain}
\end{figure}

The signature exists and is substantial. Across $354$ cells with at least $20$ false agreements to score, the mass-mean detector achieves a mean AUROC of $0.824$, with all but one cell outperforming chance and their permutation controls. Our strongest operating point occurs here: on Geometry of Truth, reviewing the top-ranked $10\%$ of claims where both probes say true recovers (recall) $77.8\%$ of the false agreements among them. Across datasets, $31$--$48\%$ of the claims the detector puts in its top $10\%$ are false agreements, against $5$--$18\%$ for random review (Figure~\ref{fig:indomain}A). RTE leaves no cell with twenty positives in its own test split and is omitted from this comparison.

The signal is not joint. Fitted on one model's activations alone, the same detector does as well: the better single model beats the joint detector by $0.012$ $[0.007, 0.017]$ AUROC with mass-mean probes and $0.014$ $[0.007, 0.021]$ with logistic probes, and the sum of the two single-model scores matches it within $0.01$. The joint detector beats only the weaker model alone ($+0.06$ and $+0.08$). Shared errors are detectable because each model signals its own errors, not through structure that appears only when both are read together. Nor is the signal specific to agreement: ranking the same claims by the two models' logistic truth probes, trained only on truth labels, reaches $0.90$, above the detector by $0.07$ $[0.06, 0.09]$, and the detector's own features trained on truth reach $0.89$. A weak reader's shared errors are found best by a stronger reader.

Variation across datasets lies in transferability rather than existence. Geometry of Truth and IMDB score closely in-domain ($0.903$ and $0.867$), yet IMDB retains $89\%$ of its lift above chance when transferred, whereas Geometry of Truth retains $45\%$ (falling from $0.903$ to $0.682$, and from $78\%$ to $38\%$ of recovered false agreements at the same review budget). BoolQ and TruthfulQA retain even less ($33\%$ and $14\%$). Detectability and transferability are distinct properties; a detector that succeeds where fitted does not guarantee general utility.

The two probe families diverge in-domain, contrary to what their accuracy suggests. Paired on cells both can score, the logistic detector sits $0.074$ AUROC below the mass-mean detector ($95\%$ model-bootstrap interval $[-0.103, -0.046]$, over the $230$ cells both detectors can score; Table~\ref{tab:unsup} instead uses the $233$ logistic cells every score can rank), winning on $17\%$ of them. The probe that reads belief more accurately is the weaker in-domain detector of shared error, although it transfers more robustly ($0.684$ vs. $0.648$ held out). This is partly mechanical: a more accurate probe produces fewer false agreements, meaning only $47\%$ of logistic cells clear the twenty-positive threshold compared to $71\%$ of mass-mean cells, leaving only the hardest positives behind. Even on the shared cells the two detectors see different numbers of positives, so the gap mixes probe type with sample size; we report both probes rather than selecting one.

\paragraph{Transfer to unseen datasets.}
We test generalizability to unseen distributions using the leave-one-dataset-out protocol of Section~\ref{sec:method:protocol}, with the review precision in Figure~\ref{fig:indomain}B. Both detectors reach an AUROC of $0.83$ on IMDB; logistic achieves $0.78$ and mass-mean $0.68$ on Geometry of Truth, while both remain between $0.54$ and $0.68$ on BoolQ, RTE, and TruthfulQA. Transfer is uneven, and in some cells it inverts: $67$ of $428$ mass-mean cells and $30$ of $444$ logistic cells score below chance, reaching $0.11$, so a transferred detector can actively mislead. Because leave-one-dataset-out also refits the probes, probe transfer and detector transfer are confounded here. The worst cells are all Geometry of Truth held out, where the direction fitted elsewhere points the wrong way. The flagged claims are false agreements $17$--$61\%$ of the time against base rates of $10$--$36\%$; on TruthfulQA the detector barely beats random review ($40\%$ against $36\%$).

Cells with fewer than $20$ false agreements in the held-out split are excluded because AUROC is uninformative on sparse positives. This removes $69$ of $105$ cells for RTE under mass-mean and $54$ of $105$ under logistic probes, making RTE's transferred bar the least supported in Figure~\ref{fig:indomain}; no other dataset loses more than one cell.

The operational operating point highlights this limitation. Reviewing the top-ranked $10\%$ recovers $11$--$39\%$ of false agreements with mass-mean probes and $14$--$48\%$ with logistic probes, tracking ranking quality closely. Where the signature transfers poorly, it captures little of the blind spot.

\subsection{What the false-agreement detector adds over a free alternative}
\label{sec:results:unsup}

Table~\ref{tab:unsup} separates three deployment regimes: no labels at all (the scores of Section~\ref{sec:method:unsup}), labels from related datasets only (the transferred detector), and labels on the target distribution (the in-domain detector). Only the first matches the oversight setting.

\input{tables/unsup.tex}

The anomaly score is inconsistent across datasets: although it beats chance in $81\%$ and $84\%$ of cells, its aggregate is driven by Geometry of Truth ($0.889$ and $0.866$) and falls to $0.496$ and $0.526$ on BoolQ. Joint representational unusualness is therefore not a general signal.

The \emph{margin} score performs differently. As expected for a confidence score, it beats chance in every cell under both probes, outperforming the supervised detector more reliably. With logistic probes, it reaches an AUROC of $0.773$, exceeding the supervised detector fitted on the same cells with labels ($0.728$) and outperforming the transferred supervised detector ($0.657$). With mass-mean probes, the ordering reverses, and the supervised detector wins by $0.119$. Thus, the two probes disagree on the optimal method, leading us to report both. As with the detector, the second model does not help: the better probe's own margin beats the two-probe margin by $0.040$ $[0.026, 0.055]$ with mass-mean probes and $0.018$ $[0.009, 0.029]$ with logistic probes. Knowing which probe is better takes labels, though, and the two-probe margin beats the weaker probe alone by $0.06$--$0.07$, so it is the label-free default.

This disagreement is interpretable. A logistic probe is a fitted classifier, making its distance from the decision boundary a calibrated measure of certainty. A mass-mean probe is a difference of centroids, where belief is an uncalibrated projection whose magnitude carries less meaning. Compared within a probe, the labelled detector leads by $0.119$ with mass-mean probes ($0.824$ against $0.705$) and trails by $0.045$ with logistic ones ($0.728$ against $0.773$), whereas a cross-dataset supervised detector underperforms both ($0.641$ and $0.657$ on identical cells, the cells every method can score; over all $428$ transferable cells the mass-mean figure is $0.648$). At a $10\%$ review budget, the margin achieves a precision of $0.232$ against a baseline rate of $0.102$, surfacing false agreements at more than twice the rate of random review.

In short, labels provide value only when sampled from the target deployment distribution. External labels do not transfer well enough to beat label-free scoring, and target distributions by definition cannot be labeled during deployment.

%% file: tables/consistency_floor.tex
\begin{table}[t]
\centering
\footnotesize
\setlength{\tabcolsep}{3pt}
\begin{tabular}{llrrrrrrrrrrr}
\toprule
& & & & \multicolumn{6}{c}{median over model pairs} & \multicolumn{3}{c}{model pairs where} \\
\cmidrule(lr){5-10}\cmidrule(lr){11-13}
Dataset & Reader & $M$ & $N$ & Fr$_\mathrm{lo}$ & CF & $p_1p_2$ & FA & Fr$_\mathrm{hi}$ & $C$ & CF$>$0 & CF/FA & CF$>$$p_1p_2$ \\
\midrule
GoT & Output & $15$ & $3830$ & $0.000$ & $\mathbf{0.015}$ & $0.007$ & $0.016$ & $0.016$ & $0.03$ & $79/105$ & $0.98$ & $38/105$ \\
 & Mass-mean & $15$ & $3830$ & $0.000$ & $\mathbf{0.001}$ & $0.001$ & $0.002$ & $0.007$ & $0.63$ & $94/105$ & $0.77$ & $75/105$ \\
 & Logistic & $15$ & $3830$ & $0.000$ & $\mathbf{0.000}$ & $0.000$ & $0.000$ & $0.001$ & $0.88$ & $22/105$ & $0.48$ & $21/105$ \\
\midrule
RTE & Output & $15$ & $277$ & $0.173$ & $\mathbf{0.206}$ & $0.283$ & $0.329$ & $0.408$ & $0.16$ & $105/105$ & $0.71$ & $7/105$ \\
 & Mass-mean & $15$ & $277$ & $0.000$ & $\mathbf{0.014}$ & $0.090$ & $0.123$ & $0.238$ & $0.48$ & $88/105$ & $0.21$ & $0/105$ \\
 & Logistic & $15$ & $277$ & $0.000$ & $\mathbf{0.029}$ & $0.056$ & $0.083$ & $0.162$ & $0.51$ & $105/105$ & $0.39$ & $7/105$ \\
\midrule
BoolQ & Output & $14$ & $3258$ & $0.132$ & $\mathbf{0.265}$ & $0.310$ & $0.355$ & $0.482$ & $0.22$ & $91/91$ & $0.74$ & $4/91$ \\
 & Mass-mean & $14$ & $3258$ & $0.000$ & $\mathbf{0.003}$ & $0.063$ & $0.150$ & $0.166$ & $0.35$ & $70/91$ & $0.05$ & $4/91$ \\
 & Logistic & $14$ & $3258$ & $0.000$ & $\mathbf{0.029}$ & $0.049$ & $0.081$ & $0.170$ & $0.55$ & $91/91$ & $0.37$ & $0/91$ \\
\midrule
IMDB & Output & $13$ & $5000$ & $0.093$ & $\mathbf{0.120}$ & $0.124$ & $0.163$ & $0.168$ & $0.01$ & $66/78$ & $0.85$ & $14/78$ \\
 & Mass-mean & $13$ & $5000$ & $0.000$ & $\mathbf{0.018}$ & $0.031$ & $0.068$ & $0.116$ & $0.47$ & $78/78$ & $0.34$ & $16/78$ \\
 & Logistic & $13$ & $5000$ & $0.000$ & $\mathbf{0.015}$ & $0.013$ & $0.042$ & $0.093$ & $0.54$ & $78/78$ & $0.37$ & $59/78$ \\
\bottomrule
\end{tabular}
\caption{The consistency floor (CF) against every other bound on false agreement (FA), at $\tau=0.5$. Each of $N$ contradiction pairs holds one true and one false statement; CF is the rate at which both readers accept both halves, so it needs no labels and is a lower bound on FA with the same denominator. $M$ models give $\binom{M}{2}$ model pairs. FA, the independence prediction $p_1p_2$ and both Fr\'echet bounds (Fr$_\mathrm{lo}$, Fr$_\mathrm{hi}$) use labels; $C$ is the coverage of Equation~\ref{eq:coverage}, the share of the better reader's errors that disagreement exposes. CF$>$0 counts pairs whose 95\% Clopper--Pearson lower limit is positive; CF/FA is pooled over pairs; CF$>$$p_1p_2$ counts pairs where the label-free floor alone exceeds what independent errors would produce. Models without complements for a dataset are absent, so $M<15$ on BoolQ and IMDB and the table covers fewer cells than the $498$ of the main grid.}
\label{tab:consistency-floor}
\end{table}

%% file: tables/unsup.tex
\begin{table}[h]
\centering
\small
\begin{tabular}{llrrrr}
\toprule
& & \multicolumn{2}{c}{mass-mean probes} & \multicolumn{2}{c}{logistic probes} \\
\cmidrule(lr){3-4}\cmidrule(lr){5-6}
Method & Labels needed & AUROC & $>$ chance & AUROC & $>$ chance \\
\midrule
Supervised, fitted in-domain & on the target & $0.824$ & $100\%$ & $0.728$ & $99\%$ \\
Supervised, transferred & other datasets & $0.641$ & n/a & $0.657$ & n/a \\
\addlinespace
Label-free: margin & none & $0.705$ & $100\%$ & $\mathbf{0.773}$ & $100\%$ \\
Label-free: anomaly & none & $0.626$ & $81\%$ & $0.571$ & $84\%$ \\
\bottomrule
\end{tabular}
\caption{Supervised and label-free detection AUROC. Mean AUROC is computed over cells scorable by every method ($354$ for mass-mean; $233$ for logistic). ``$>$ chance'' reports the share of cells with AUROC above $0.5$, indicating how broadly performance exceeds chance.}
\label{tab:unsup}
\end{table}

%% file: sections/discussion.tex
\section{Discussion and Limitations}
\label{sec:discussion}

\paragraph{What the floor certifies and what it cannot.}
The consistency floor turns contradictions into a certified minimum of shared error, computable on deployment data without labels. It is only a floor. A shared flip is internally consistent, so no contradiction check can tell it from agreement on the truth, and the gap to false agreement stays invisible without labels. Nor can it be closed from above: the only label-free ceiling is the agreement rate itself (Proposition~\ref{prop:sharp}), and agreement-based estimators read shared errors as correct answers. A pipeline with complements can already filter self-contradictions, so the floor's use is as an audit statistic: it measures, before deployment and without labels, how much of two readers' shared error such a filter would remove, and Corollary~\ref{cor:flip} says nothing can be certified about the rest.

How much the floor recovers also depends on reader design and calibration. Yes-biased outputs often accept both claims of a pair, so the floor catches most of their shared error; each catch is a real shared error, but the high rate partly reflects response bias. Removing both design choices narrows the gap: rate-matched outputs fall to $30$--$78\%$, and probes trained on one claim rise to $20$--$98\%$ (Table~\ref{tab:reader-design}).

\paragraph{Implications for overseer selection.}
Accuracy alone does not decide whether a second overseer helps; what matters is how its errors overlap with the first. Reading activations instead of outputs lowers false agreement by making each reader more accurate, not more independent, and model family is an unreliable proxy for independence: cross-family pairs have less-correlated errors, reliably so ($0.36$ to $0.26$ with mass-mean probes, $0.42$ to $0.35$ with logistic ones), but correlation governs only about half of false agreement, so the reduction is worth roughly one percentage point against pair-to-pair variation spanning $0.5\%$ to $29\%$. Overseers should be chosen by measuring both accuracy and shared error on deployment-like data, with the floor as a label-free lower bound on the latter.

\paragraph{Why false agreement is high.}
Three effects combine. First, accuracy: the independence term $p_1p_2$ accounts for about half of false agreement, so two reasonably accurate readers share many errors even if their errors are uncorrelated. Second, response bias: yes-biased outputs push $p_1+p_2$ above $1$, which forces overlap (a positive Fr\'echet lower bound). Third, shared difficulty: errors likely concentrate on claims that are hard for every model, such as popular misconceptions or misleading surface cues, which would explain why changing model family barely helps.

\paragraph{Comparison with prior output-level measurements.}
For scale, \citet{kim2025correlated} report model pairs sharing the same wrong output on $9.2\%$ of HELM and $17.7\%$ of Hugging Face leaderboard questions; our logistic probes, counted over all claims, share errors on $6.2\%$. These figures use different denominators, models and tasks, so they are not directly comparable.

\paragraph{The residual blind spot.}
The floor bounds how much shared error there is; detection baselines ask which agreed claims hide it; a stronger truth probe does best. False agreement is detectable in-domain, but from either model's activations alone; reading both adds nothing. Uneven transfer argues against a universal detector reused without validation. A practical alternative is triage: disagreement routes claims for review, and a score needing no labels beyond those that fit the probes ranks the agreed ones. The smaller of the two probes' margins beats an in-domain detector with logistic probes ($0.77$ against $0.73$), trails it with mass-mean probes ($0.71$ against $0.82$), and beats transferred detectors with both, so supervised detection pays off only where target labels already exist.

\paragraph{Weaker readouts fail together more, but more legibly.}
With CCS probes and base models (Appendix~\ref{app:ccs-base}), weaker readouts produce more shared error, but more systematic and so easier to rank. Detectability thus shifts with readout accuracy, and a detector validated on strong probes should not be assumed to hold for weak ones.

\paragraph{Scope and limitations.}
We study the verdicts of two readers, single-token outputs and supervised linear probes, not beliefs; a false agreement may reflect a shared misconception, a shared reader failure, or both. The comparison is asymmetric, since probes are trained on labels and outputs are read zero-shot, and both probes in a pair share the same labels and split, a source of shared error outputs lack. Only the Geometry of Truth complements are logical negations; the rest are templates, so the floor depends on phrasing. In deployment, complements must be generated, and negation is a known failure mode for probes \citep{levinstein2024still,marks2024geometry}; the floor is guaranteed only where each pair holds exactly one true claim, so a malformed complement, with zero or two true claims, can inflate it. Claims come from fixed datasets rather than the overseen model's own generations, where errors may correlate more; that setting remains to be tested. Cells reuse models and items, so they are not independent observations. In-domain detection needs at least $20$ false agreements per cell, which excludes $144$ of $498$ cells with mass-mean probes, $265$ with logistic probes, and all of RTE, whose small size also gives single-cell rates a standard error near $5$ points. These limits narrow the claim but not the finding: agreement between two readers, of outputs or activations, can hide substantial shared error, of which a label-free floor certifies only part. Extending the floor to more than two readers, where every reader accepting both claims certifies an error shared by all, is a natural next step.

%% file: sections/conclusion.tex
\section{Conclusion}
\label{sec:conclusion}

Agreement between two overseers is useful only to the extent that their errors
differ, and measuring how far they differ has required labels that oversight
does not have. The consistency floor removes that requirement for part of the
problem. Pairing each claim with a complement certifies a minimum of shared
error without labels, and in most model pairs that minimum is positive. Reading
activations instead of outputs lowers false agreement, but through accuracy
rather than independence, and leaves shared errors that are more often
consistent and so harder to certify. A false-agreement detector can rank the
remainder within a domain, but its signature does not transfer uniformly to
unseen data distributions. Cross-model oversight should therefore measure
shared error directly, with the floor where labels are absent, rather than
infer independence from model family or accuracy alone.

%% file: sections/appendix_floor.tex
\section{The Consistency Floor Across Datasets and Readers, and Label-Free Ceilings}
\label{app:floor}

\begin{figure}[t]
\centering
\includegraphics[width=\textwidth]{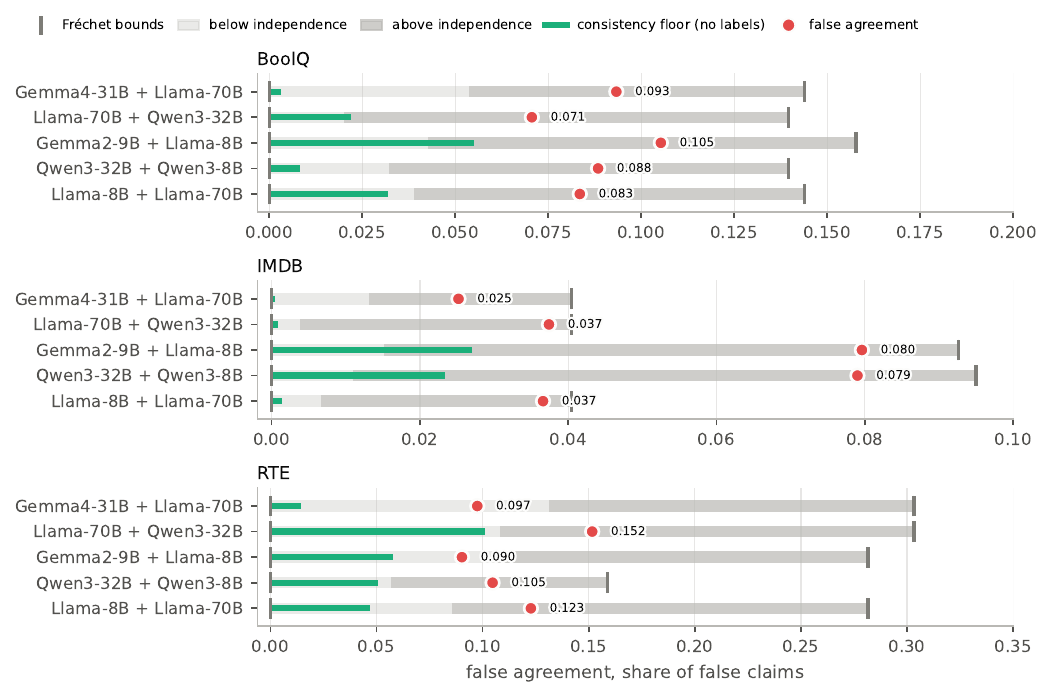}
\caption{The same five model pairs as Figure~\ref{fig:bound}, read by
mass-mean probes. Gray bars and ticks are Fr\'echet intervals, pale below
independence and darker above it; the green segment from zero is the
consistency floor and the red dot is measured false agreement, all on the same
complement pairs ($\tau=0.5$). Each panel has its own scale.}
\label{fig:bound-probe}
\end{figure}

Figure~\ref{fig:bound-probe} repeats Figure~\ref{fig:bound} with mass-mean
probes in place of the outputs: the same pairs, datasets and complement pairs,
with probes cross-fitted over pairs as described in
Section~\ref{sec:method:floor}. Probe false agreement sits lower in its
Fr\'echet interval than output false agreement, and the floor recovers a smaller
share of it, most visibly on BoolQ and RTE.

\begin{figure}[p]
\centering
\includegraphics[width=\textwidth]{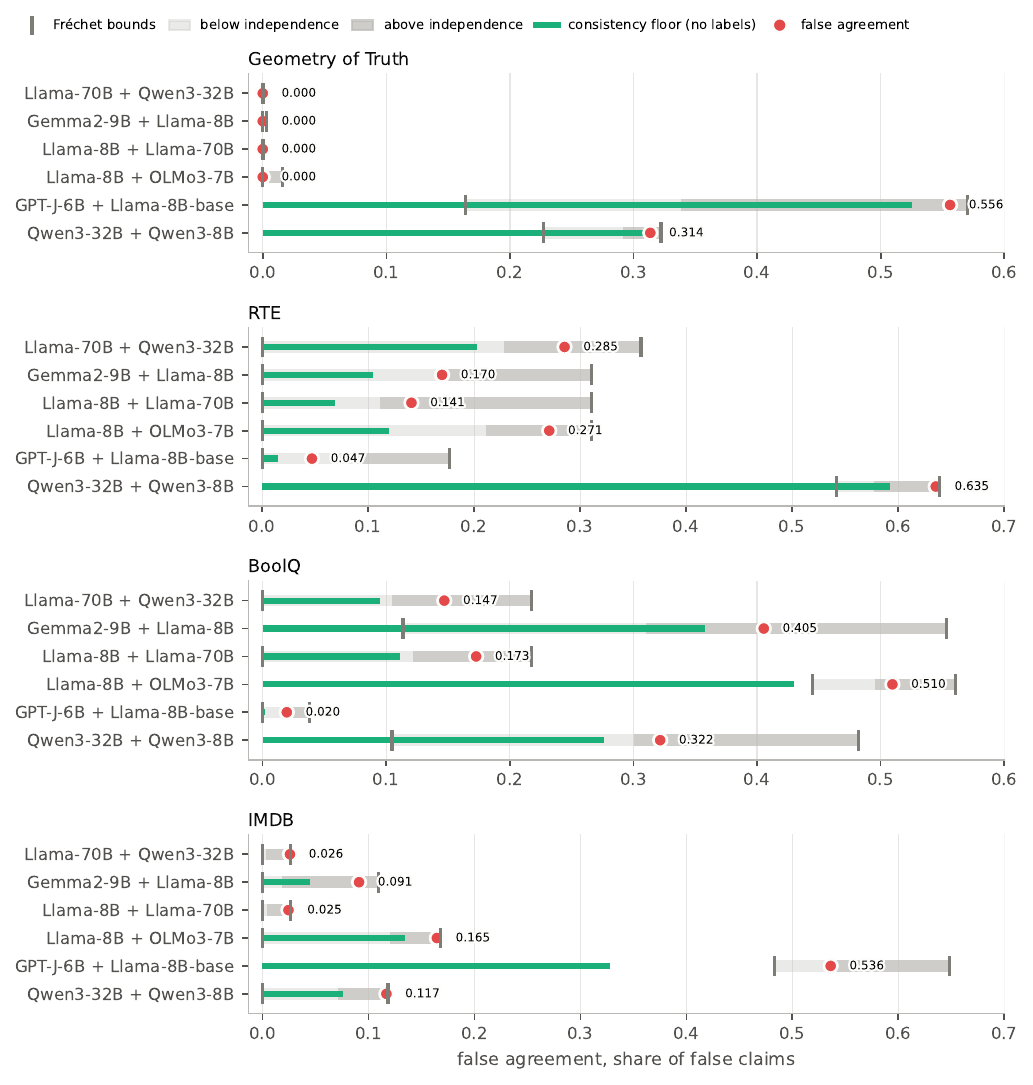}
\caption{The consistency floor on the four datasets with label-free complements, six model pairs, model
outputs. Gray bars and ticks are Fr\'echet intervals, pale below independence
and darker above it; the green segment from zero is the consistency floor and
the red dot is measured false agreement, all on the same complement pairs
($\tau=0.5$). Each panel has its own scale.}
\label{fig:bound-all}
\end{figure}

Figure~\ref{fig:bound-all} then extends Figure~\ref{fig:bound} to all four datasets with label-free complements
with six model pairs, read from the models' outputs. Only pairs whose two models
have complements on every dataset qualify, and pairs containing a model whose
output answers \textsc{yes} to nearly every claim are left out, since their
Fr\'echet interval collapses to a point. Two things are visible here that the
main figure does not show. On Geometry of Truth most output pairs make no shared
error at all; the exceptions are pairs with a strongly \textsc{yes}-biased model.
And the floor can sit below the Fr\'echet lower bound (BoolQ, Llama-8B +
OLMo3-7B; IMDB, GPT-J-6B + Llama-8B-base): the two bounds use different
information, the marginals on one side and contradictions on the other, and
neither implies the other.

\begin{figure}[h]
\centering
\includegraphics[width=\textwidth]{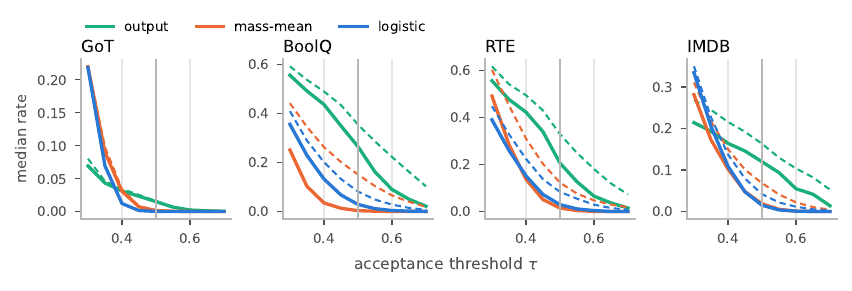}
\caption{Median consistency floor (solid) and false agreement (dashed) over model pairs, against the acceptance threshold $\tau$. Gray line: $\tau=0.5$.}
\label{fig:floor-tau}
\end{figure}

Figure~\ref{fig:floor-tau} varies the acceptance threshold. At $\tau=0.5$ each probe sits at its fitted decision boundary (Section~\ref{sec:method:probe}); raising $\tau$ shrinks the floor and false agreement together, and probe floors approach zero just above $0.5$.

Table~\ref{tab:reader-design} gives the reader-design controls of Section~\ref{sec:results:floor} per dataset.

\input{tables/reader_design.tex}

\subsection{Agreement-based ceiling baseline}
\label{app:floor:ds}

As a label-free comparison for the consistency floor, we estimate each reader's
error rate with the Dawid--Skene estimator \citep{dawid1979maximum}, aggregating
over model agreement patterns, and use $\min(\hat{p}_1,\hat{p}_2)$ as an
estimated ceiling on false agreement. This estimate relies on conditional
independence of reader errors given the label. Shared errors violate the
intuition behind that assumption: agreement on a false claim can be interpreted
as evidence of correctness, so the estimated ceiling can fall below measured
false agreement. We report this baseline to show that agreement-based estimates
do not replace labels for quantifying shared error. Table~\ref{tab:ds-ceiling}
reports the ceiling estimate and the number of model pairs for which measured
false agreement exceeds it. Probe estimates use a separate sweep split, as
noted in the table caption.

\input{tables/ds_ceiling.tex}

%% file: tables/reader_design.tex
\begin{table}[h]
\centering
\small
\begin{tabular}{lcccc}
\toprule
Reader & GoT & BoolQ & RTE & IMDB \\
\midrule
Output, own threshold & 98 / 0 & 74 / 5 & 71 / 4 & 85 / 1 \\
Output, matched to mass-mean & 78 / 1 & 33 / 29 & 30 / 27 & 45 / 17 \\
Output, matched to logistic & 78 / 1 & 36 / 25 & 32 / 23 & 37 / 22 \\
\addlinespace
Mass-mean, both claims & 77 / 1 & 5 / 73 & 21 / 47 & 34 / 26 \\
Mass-mean, one claim & 81 / 1 & 20 / 50 & 46 / 24 & 32 / 31 \\
\addlinespace
Logistic, both claims & 48 / 18 & 37 / 25 & 39 / 23 & 37 / 30 \\
Logistic, one claim & 98 / 0 & 62 / 12 & 60 / 16 & 59 / 17 \\
\bottomrule
\end{tabular}
\caption{Share of false agreement the consistency floor certifies / share that are shared flips (\%), pooled over model pairs at $\tau=0.5$. \emph{Matched} outputs are thresholded to the named probe's positive rate; \emph{one claim} probes are fitted on the original claim of each pair only, never seeing a complement.}
\label{tab:reader-design}
\end{table}

%% file: tables/ds_ceiling.tex
\begin{table}[h]
\centering
\small
\begin{tabular}{llrr}
\toprule
Dataset & Reader & DS ceiling (median) & FA$>$DS \\
\midrule
GoT & Output & $0.016$ & $9/105$ \\
 & Mass-mean & $0.009$ & $10/105$ \\
 & Logistic & $0.006$ & $0/105$ \\
RTE & Output & $0.392$ & $39/105$ \\
 & Mass-mean & $0.148$ & $40/105$ \\
 & Logistic & $0.158$ & $23/105$ \\
BoolQ & Output & $0.425$ & $33/91$ \\
 & Mass-mean & $0.173$ & $27/91$ \\
 & Logistic & $0.171$ & $0/91$ \\
IMDB & Output & $0.079$ & $51/78$ \\
 & Mass-mean & $0.034$ & $53/78$ \\
 & Logistic & $0.037$ & $51/78$ \\
\bottomrule
\end{tabular}
\caption{The label-free Dawid--Skene ceiling estimate on false agreement (all models pooled), and the model pairs whose measured FA exceeds it, i.e.\ where the ceiling failed. Output rows use the complement pairs of Table~\ref{tab:consistency-floor}; probe rows use the sweep's grid split and its probes, so they are not directly comparable with that table.}
\label{tab:ds-ceiling}
\end{table}

%% file: sections/appendix_datasets.tex
\section{Datasets and Methodology}
\label{app:datasets}

\subsection{The five datasets}

Every dataset is reduced to a set of standalone declarative claims with a binary
truth label, so that a single contrast pair can be built for each. Where the
source is a question-answering or entailment task, the reformulation is stated
below. Table~\ref{tab:datasets} summarizes the result.

\begin{table}[h]
\centering
\small
\begin{tabular}{llrrl}
\toprule
Dataset & Source & Items & Median tokens & Claim form \\
\midrule
Geometry of Truth & \citet{marks2024geometry} & $5030$ & $11$ & curated statement \\
TruthfulQA        & \citet{lin2022truthfulqa} & $1634$ & $22$ & question $+$ candidate answer \\
RTE               & \citet{wang2018glue}      & $277$  & $58$ & premise $+$ hypothesis \\
BoolQ             & \citet{clark2019boolq}    & $3270$ & $121$ & passage $+$ yes/no question \\
IMDB              & \citet{maas2011imdb}      & $5000$ & $187$ & review $+$ sentiment claim \\
\bottomrule
\end{tabular}
\caption{The five datasets after reduction to labelled claims. Token counts are
GPT-2 tokens of the claim text, before the contrast-pair suffix is appended.
IMDB is capped at $5000$ items, balanced across labels; the others are used in
full.}
\label{tab:datasets}
\end{table}

\paragraph{Geometry of Truth.}
We use four of the fifteen topic files released with
\citet{marks2024geometry}: \texttt{cities}, \texttt{sp\_en\_trans},
\texttt{larger\_than} and \texttt{companies\_true\_false}, read from the
authors' repository. These are short affirmative statements of a single fact,
for example \emph{``The city of Krasnodar is in Russia.''} or \emph{``Fifty-one
is larger than fifty-two.''}

Three further files, the negated counterparts of three of the four we use
(\texttt{neg\_cities}, \texttt{neg\_sp\_en\_trans}, \texttt{smaller\_than}),
supply the complements for the consistency floor
(Section~\ref{sec:method:floor}), matched to their affirmative twins by the
entities they name; \texttt{companies\_true\_false} has no negated file and so
contributes no complement pairs. They enter only the floor, not the probe or
detector training items. The remaining files we do not use include the logical
compositions
(\texttt{cities\_cities\_conj}, \texttt{cities\_cities\_disj}), and the
uncurated sets (\texttt{common\_claim\_true\_false},
\texttt{counterfact\_true\_false}). Our subset is therefore the affirmative,
templated portion of the benchmark, which is reflected in its statistics: it has
the shortest claims of the five datasets by a wide margin and the least
variation in length.

This matters for interpreting the results, because Geometry of Truth is the
dataset on which in-domain detection is strongest
(Section~\ref{sec:indomain}), although IMDB transfers better. Whether that reflects the factual-recall content
of the claims or their surface uniformity can be tested directly, because
every false agreement on Geometry of Truth falls inside one of its four topic
files. Refitting the transferred detector within \texttt{companies\_true\_false}
alone, where template, length and surface form are held constant, costs
$0.015$ AUROC ($0.729$ to $0.714$ over $99$ cells; $0.840$ to $0.797$ under mean
pooling). The detector is therefore reading shared error rather than topic. Using the negated files as items in their own right, not only as
complements, would separate the two, since they hold the same facts in a harder
syntactic form, and we flag this as the most informative single extension to
the data.

\paragraph{TruthfulQA.}
Each question is paired with one candidate answer to form a claim, labelled by
whether that answer is among the correct ones. Questions are kept together under
a shared group id so that a question's answers cannot be split across the train
and test halves.

\paragraph{RTE.}
The premise and hypothesis are concatenated into a single claim, labelled by
whether the premise entails the hypothesis. It is the smallest dataset at $277$
items, which is why its cells fall below the twenty-positive threshold of the
in-domain detection analysis (Section~\ref{sec:indomain}).

\paragraph{BoolQ.}
The passage and its yes/no question are concatenated, labelled by the answer.
Claims are long and highly variable in length.

\paragraph{IMDB.}
A review is paired with a claim about its sentiment, labelled by whether the
claim matches the review's label. Capped at $5000$ items with equal numbers of
each label; the other four datasets are used in full.

\subsection{Splitting and group integrity}

Every dataset is split once into train and test halves using seed $0$, and the
same split is reused by every model, so that a pair of models is always compared
on identical items. Items sharing a group id, such as the answers to one
TruthfulQA question, are assigned to the same half, so that a probe cannot
be fitted on one answer to a question and scored on another. Truncation to a
cap is stratified by label and respects groups. Exact Hugging Face checkpoint
repository IDs for all evaluated models are listed in the supplementary
archive in \texttt{model\_manifest.csv}.

\subsection{Thresholds}

False agreement is a rare event on some datasets, and a rate estimated from a
handful of items is not informative. A threshold is therefore applied to the
detection results and stated wherever it affects a number. Cells with fewer than $20$ false
agreements in the evaluated split are excluded from detection results; this
removes $144$ of $498$ cells in the in-domain analysis under mass-mean probes
and $265$ under logistic ones, falling hardest on RTE, which retains none: its
test split is too small to leave twenty false agreements in any cell, so RTE
appears in the transfer results but not the in-domain ones. The shortage is specific to RTE as an in-domain fit, since
its test split holds a median of $62$ false claims per cell; held out, it is
scored by probes fitted on the other four datasets, and $36$ (mass-mean) and $51$ (logistic) of its $105$ cells clear the threshold. No threshold is applied to the false-agreement rates themselves,
which are averaged over all cells.

The false-agreement rate and the detector's positives count different things,
and the counts are not interconvertible. A false-agreement rate is measured on the dataset's own test
split with each probe at the threshold it was fitted with; the detector is
fitted on other datasets and applied to the whole held-out dataset at a
recalibrated threshold. A cell can therefore report a rate of under $1\%$ and
still give the detector hundreds of positives to score, and the number of false
agreements available to the detector cannot be recovered from the rate column
of any table.

\subsection{Extraction and reasoning models}

Six of the fifteen models are hybrid thinking models, spanning three of the five
families. We extract with no chat template and thinking off, so no reasoning
trace is generated. This keeps the comparison identical across reasoning and
non-reasoning models, and it keeps the output reader a genuine
single-token Yes/No logprob, which it would not be if the next token began a
trace. The probe therefore reads pre-reasoning belief, which is a narrower
object than belief, and the paper says so wherever the distinction matters.

\subsection{Label-free scores}
\label{app:labelfree}

Three label regimes have to be kept apart, because they license different
claims. A detector fitted and scored inside one dataset needs labels \emph{on
the distribution being overseen}, which is exactly what the oversight setting
denies: if the claims could be checked, the problem would not arise. The
leave-one-dataset-out protocol removes that requirement, fitting the detector
on other datasets and applying it to an unlabelled target, so labels are still
needed but elsewhere. The third regime needs none, and is what the label-free
scores test. In all three the probes stay supervised: each model's
probe is fitted once, on whatever labelled data exists, at a cost paid per model
rather than per deployment. The question is only whether the false-agreement detector can be had for free given two probes that already exist.

\paragraph{Margin.} $-\min(|b_1 - \tfrac12|,\, |b_2 - \tfrac12|)$, where $b_i$
is probe $i$'s belief: how weakly the more hesitant of the two agreed. It
requires no fitting of any kind, so a system that has the probes already has
this score.

\paragraph{Anomaly.} Mahalanobis distance from the centre of the both-agree
pool, in a basis of its top $32$ principal components, or fewer where the pool
has lower rank. The projection and covariance are estimated on the unlabelled
pool itself. It tests whether shared mistakes occupy unusual regions of the
joint representation rather than typical ones, and unlike the margin it reads
both models' activations rather than only the probes' scalar outputs.

\paragraph{Why not CCS.} Its objective is defined over a claim and its negation,
and the quantity to be separated at this stage is not a claim's truth but
whether an \emph{agreement} is mistaken. Constructing the corresponding
contrastive pair would require a second extraction over prompts about the
agreement itself, which we do not have. CCS appears in this paper only as an
alternative probe reader.

\paragraph{A caveat on reporting.} Choosing the better of the two scores per
cell would itself consume labels, so that quantity would be an upper bound
rather than a method. We report each score separately.

%% file: sections/appendix_ccs.tex
\section{Fitting CCS Without Reading the Token Identity}
\label{app:ccs}

Two implementation choices decide whether a CCS probe reads truth or nothing at
all. Both failures are silent: the fit converges, the loss is low, and the
probe scores at chance.

\paragraph{Standardize each half of the contrast pair separately.}
The two halves of a contrast pair differ by a single token, so their difference
carries a large constant direction, the token identity, shared by every item.
Measured on Geometry of Truth, its magnitude is a median of $1.7\times$ the
item-to-item variation across our models, and up to $4.2\times$. A shared
normalizer leaves that direction in place, and CCS can then satisfy both of its
loss terms by reading it alone: $p^+ \approx 1$ and $p^- \approx 0$ on every
item is perfectly consistent, perfectly confident, and carries no information
about the claim. Subtracting each half's own mean removes the direction, which
forces the consistency term to be satisfied by something item-specific
\citep{burns2023ccs}.

\paragraph{Rank restarts by belief spread, not by training loss.}
For the same reason, the minimum-loss restart is the wrong one to keep. The
degenerate solution attains a lower loss ($\sim\!10^{-4}$) than any probe that
tracks truth ($\sim\!10^{-2}$), so selecting on loss selects the uninformative
probe, and does so while looking healthy. We instead reject restarts whose
belief output has near-zero spread across items, which is a label-free
criterion, and rank only the survivors. Without this filter we observed a single
cell swing between $0.43$ and $0.98$ AUROC across splits.

%% file: sections/appendix_gate.tex
\section{Do the Probes Read Truth, or Confidence?}
\label{app:gate}

A probe that merely re-encoded the model's stated confidence would add nothing
over the output reader. We test for that directly.
For each model and dataset we take the half of items the model is most
confident about, measured by how far its single-token $p(\textsc{yes})$ sits
from $0.5$, and score two readers against ground truth on that half: the probe, and
the model's own signed confidence. The \emph{edge} is the first AUROC minus the
second. A positive edge means the probe carries information the output does
not.

The confident half is the demanding place to ask. It is where the model's own
answer is the strongest available predictor, and it is the half oversight
actually cares about, since a model that is confident and wrong is what passes
an output-level check unchallenged.

\input{tables/gate_a.tex}

Table~\ref{tab:gate-a} gives the result in full, and it is uneven. On
TruthfulQA every model passes under both probe estimators, with median edges of
$+0.18$ and $+0.30$; that dataset answers $39\%$ of its confident items wrongly,
so there is a large pool of confidently-wrong claims for a probe to separate.
On Geometry of Truth the logistic probe passes for all fifteen models and the
mass-mean probe for eleven.

Elsewhere the edge is small. Median edges on IMDB, BoolQ and RTE are $+0.008$,
$+0.021$ and $+0.051$ for the mass-mean probe, positive for $12$ of $13$, $12$
of $15$ and $11$ of $15$ models, and $+0.073$, $+0.116$ and $+0.019$ for the
logistic probe. On those datasets the models are usually right when they are
confident, $74\%$ to $83\%$ of the time, so output confidence is
already a good predictor and there is little left for a probe to add. We read
this as a limit on the claim rather than against it: the probes earn their place
where the model is confidently wrong often enough for the distinction to matter,
and that is the regime the oversight question is about.

Two caveats. The comparison is between one probe and one model's output, so it
speaks to whether a single readout is informative, not to how two readouts
share errors, which is the subject of the rest of the paper. And it is an
AUROC comparison, so it concerns ranking rather than the thresholded verdicts
from which false agreement is computed.

%% file: tables/gate_a.tex
\begin{table}[!htbp]
\centering
\footnotesize
\setlength{\tabcolsep}{4pt}
\caption{Probe versus output confidence by model and dataset: the probe's AUROC minus the model's own signed-confidence AUROC, on the half of items the model is most confident about. Each cell gives mass-mean / logistic. A positive value means the probe reads something the output does not. The last row is the share of those confident items the model answers wrongly, which is how much room the gate leaves to win. Two models have no IMDB extraction.}
\label{tab:gate-a}
\begin{tabular}{lccccc}
\toprule
Model & GoT & IMDB & BoolQ & TruthfulQA & RTE \\
\midrule
GPT-J-6B & $+0.23$ / $+0.54$ & $+0.46$ / $+0.50$ & $+0.14$ / $+0.21$ & $+0.20$ / $+0.31$ & $+0.25$ / $+0.28$ \\
Gemma2-9B & $+0.01$ / $+0.07$ & $+0.01$ / $+0.01$ & $+0.04$ / $+0.07$ & $+0.05$ / $+0.15$ & $+0.01$ / $-0.00$ \\
Gemma4-12B-base & $+0.07$ / $+0.14$ & $+0.28$ / $+0.41$ & $-0.00$ / $+0.10$ & $+0.07$ / $+0.26$ & $+0.09$ / $+0.02$ \\
Gemma4-31B & $+0.28$ / $+0.45$ & n/a & $+0.08$ / $+0.15$ & $+0.36$ / $+0.44$ & $+0.08$ / $+0.01$ \\
Llama-405B & $+0.02$ / $+0.02$ & $+0.00$ / $-0.00$ & $-0.02$ / $+0.01$ & $+0.03$ / $+0.06$ & $+0.01$ / $+0.02$ \\
Llama-70B & $+0.00$ / $+0.00$ & $+0.00$ / $+0.00$ & $-0.01$ / $+0.02$ & $+0.06$ / $+0.12$ & $-0.00$ / $+0.00$ \\
Llama-70B-base & $-0.10$ / $+0.05$ & $+0.09$ / $+0.39$ & $+0.01$ / $+0.12$ & $+0.18$ / $+0.30$ & $-0.06$ / $-0.02$ \\
Llama-8B & $-0.01$ / $+0.00$ & $+0.00$ / $+0.00$ & $+0.01$ / $+0.08$ & $+0.15$ / $+0.25$ & $-0.01$ / $-0.03$ \\
Llama-8B-base & $-0.07$ / $+0.06$ & $+0.21$ / $+0.28$ & $+0.01$ / $+0.15$ & $+0.23$ / $+0.38$ & $+0.08$ / $+0.12$ \\
Olmo3-7B & $-0.00$ / $+0.02$ & $+0.11$ / $+0.16$ & $+0.06$ / $+0.16$ & $+0.25$ / $+0.35$ & $+0.05$ / $+0.04$ \\
Qwen3-1.7B & $+0.29$ / $+0.34$ & $+0.13$ / $+0.13$ & $+0.17$ / $+0.13$ & $+0.24$ / $+0.30$ & $+0.21$ / $+0.25$ \\
Qwen3-32B & $+0.03$ / $+0.13$ & $+0.00$ / $-0.00$ & $+0.02$ / $+0.02$ & $+0.04$ / $+0.17$ & $+0.12$ / $+0.14$ \\
Qwen3-8B & $+0.19$ / $+0.35$ & $-0.01$ / $-0.00$ & $+0.02$ / $+0.07$ & $+0.11$ / $+0.32$ & $+0.04$ / $-0.10$ \\
Qwen3-8B-base & $+0.17$ / $+0.33$ & $+0.01$ / $+0.07$ & $+0.05$ / $+0.17$ & $+0.23$ / $+0.38$ & $-0.08$ / $-0.22$ \\
Qwen3-27B & $+0.30$ / $+0.36$ & n/a & $+0.25$ / $+0.31$ & $+0.33$ / $+0.42$ & $+0.41$ / $+0.40$ \\
\midrule
\itshape confidently wrong & $12\%$ & $18\%$ & $17\%$ & $39\%$ & $26\%$ \\
\bottomrule
\end{tabular}
\end{table}

%% file: sections/appendix_sweeps.tex
\section{Probe Depth}
\label{sec:results:depth}

The results in Section~\ref{sec:results} read the final layer. Truth directions
are often reported as strongest in middle layers, so we repeat the measurements
at $0.5$ and $0.75$ of model depth. Only the nine locally extracted models hold
more than a final layer, so this covers $36$ pairs rather than $105$, and a pair
is dropped rather than read at two different depths.

\input{tables/depth.tex}

Table~\ref{tab:depth} reports the sweep, and depth matters little. With
$95\%$ intervals from a bootstrap over models, which respects the models shared
between cells, one of the twelve comparisons against the final layer excludes
zero: depth $0.50$ improves in-domain detection with mass-mean probes by
$0.024$ $[0.002, 0.049]$. Transfer and the label-free margin do not change
measurably at either depth, and neither does false agreement ($9.9\%$ at the
final layer, $10.1\%$ at $0.75$, $[-0.4, 0.8]$ points). Reading the final layer
therefore costs little.

\section{CCS on Base Models: a Larger, Still Detectable Blind Spot}
\label{app:ccs-base}

CCS does not reliably read truth on base models (belief AUROC
$0.478$--$0.580$, measured with mean pooling), so we do not treat it as an
unsupervised replacement for the main probes. We can still ask whether the
false-agreement detector can rank the false agreements CCS probe pairs produce.
Table~\ref{tab:ccs-base-models} covers the full grid at the last token. False
agreement rises from $17.3\%$ when neither model is a base model to $24.4\%$
when both are, while detection AUROC stays level or rises ($0.851$ to
$0.881$); per base model it ranges from $0.849$ to $0.895$, against
$0.796$--$0.895$ for instruct models. Weaker readouts therefore leave a larger
blind spot with no loss in detectability for the supervised detector; this does
not establish label-free detection. The two-base-model row has $25$ cells, so
that comparison is exploratory.

\input{tables/ccs_base_models.tex}

%% file: tables/depth.tex
\begin{table}[h]
\centering
\small
\begin{tabular}{llrrrr}
\toprule
Measurement & Probe & $d=0.50$ & $d=0.75$ & $d=1.00$ & $\Delta_{0.75}$ \\
\midrule
In-domain detection & mass-mean & $0.847$ & $0.838$ & $0.823$ & $+0.015$ \\
                    & logistic & $0.730$ & $0.737$ & $0.733$ & $+0.003$ \\
\addlinespace
Transfer (LOO)      & mass-mean & $0.610$ & $0.625$ & $0.605$ & $+0.020$ \\
                    & logistic & $0.659$ & $0.685$ & $0.679$ & $+0.006$ \\
\addlinespace
Label-free margin   & mass-mean & $0.664$ & $0.699$ & $0.695$ & $+0.004$ \\
                    & logistic & $0.746$ & $0.775$ & $0.757$ & $+0.018$ \\
\bottomrule
\end{tabular}
\caption{Detection AUROC by probe depth, last token, on the nine models that hold all three depths. Each depth is paired cell by cell against the final layer on the cells both can score (at least 20 false agreements in each); $d=1.00$ and $\Delta_{0.75}$ are taken on the cells shared with $d=0.75$, so the $d=0.50$ column is not paired with the $d=1.00$ column shown.}
\label{tab:depth}
\end{table}

%% file: tables/ccs_base_models.tex
\begin{table}[!htbp]
\centering
\small
\caption{CCS-probe pairs stratified by the number of base models in the pair. The detection AUROC is from the false-agreement detector; false agreement is the share of false claims both CCS probes classify as true. Cells need at least 20 false agreements to score, leaving $362$ of the $533$ CCS cells. The two-base-model stratum is exploratory.}
\label{tab:ccs-base-models}
\begin{tabular}{lrrr}
\toprule
Base models in pair & Cells & Detection AUROC & False agreement \\
\midrule
None & $182$ & $0.851$ & $17.3\%$ \\
One & $155$ & $0.861$ & $19.3\%$ \\
Two & $25$ & $0.881$ & $24.4\%$ \\
\bottomrule
\end{tabular}
\end{table}

%% file: sections/appendix_pooling.tex
\section{Pooling and Claim Length}
\label{app:pooling}

Every result in this paper reads the final token of each contrast half. Mean
pooling over the claim is the more common choice, and the one an embedding API
returns, but for a contrast pair it is not an alternative so much as the same
quantity divided by claim length.

Under causal attention the \texttt{Yes} and \texttt{No} halves of a contrast
pair share every prefix state: they differ only in the final token. Pooling
each half over $L$ positions and subtracting therefore gives
\[
  \overline{h}_{\textsc{yes}} - \overline{h}_{\textsc{no}}
  = \frac{h_{\textsc{yes}} - h_{\textsc{no}}}{L},
\]
the last-token difference divided by the claim's length. We confirmed this on
the cached activations: the cosine between the mean-pooled and last-token
difference is $0.99999$ (minimum $0.99988$ over $5030$ Geometry of Truth
items), and the ratio of their norms is an integer equal to the sequence
length, to a mean absolute error of $0.002$. Mean pooling contributes nothing
the final token does not, except a division by $L$.

That division matters because $L$ varies by an order of magnitude across these
datasets. Counting the whole prompt, which is what the mean runs over, the median
is $17$--$20$ tokens on Geometry of Truth and $194$--$196$ on IMDB across the
models' tokenizers; Table~\ref{tab:datasets} counts the claim alone. The contrast feature is therefore scaled by a
dataset-specific constant, which rescales each probe's decision boundary
without rescaling its training data, and the leave-one-dataset-out protocol
applies probes fitted on one length regime to another.

\paragraph{What the scaling costs.}
Measured over the full grid at both pooling choices, the effects fall on exactly
the datasets the argument predicts. Probe AUROC rises by $0.022$
with mass-mean probes and is unchanged with logistic ones, with the largest
gain on RTE, whose claims are long, and none on Geometry of Truth, whose claims
are short. False agreement falls on every dataset, from $11.0\%$ to $9.1\%$
with mass-mean probes and from $12.2\%$ to $8.8\%$ with logistic ones, and the
fall again tracks claim length: RTE drops by nine percentage points under
mass-mean probes while Geometry of Truth moves by less than one.

Two further symptoms are worth recording. The mass-mean probe's positive rate
on Geometry of Truth fell from $0.66$ to $0.55$ once the scaling was removed, so
the mean-pooled version was calling two claims in three true and reporting
false agreement over a pool inflated accordingly. And the permutation control
for the transfer detector on Geometry of Truth, which should sit at $0.5$,
averaged $0.474$ under mean pooling and $0.498$ at the final token.

\paragraph{What it leaves intact.}
No conclusion in the paper depends on the choice. False agreement remains
large and remains above its independence reference on every dataset; probe
errors remain positively correlated in the large majority of cells; model-family
diversity still lowers error correlation without measurably shrinking the blind
spot; and the blind spot is still detectable in-domain and transfers unevenly.
What changes is the magnitude of individual numbers and, on Geometry of Truth,
how well detection transfers. We report final-token results throughout and flag this as a confound any probe
reading a contrast pair should check, since it follows from causal attention
rather than from these models.

%% file: sections/appendix_deployment.tex
\section{Coverage from Disagreement and Detector Review}
\label{app:deployment-coverage}

\begin{figure}[t]
\centering
\includegraphics[width=\textwidth]{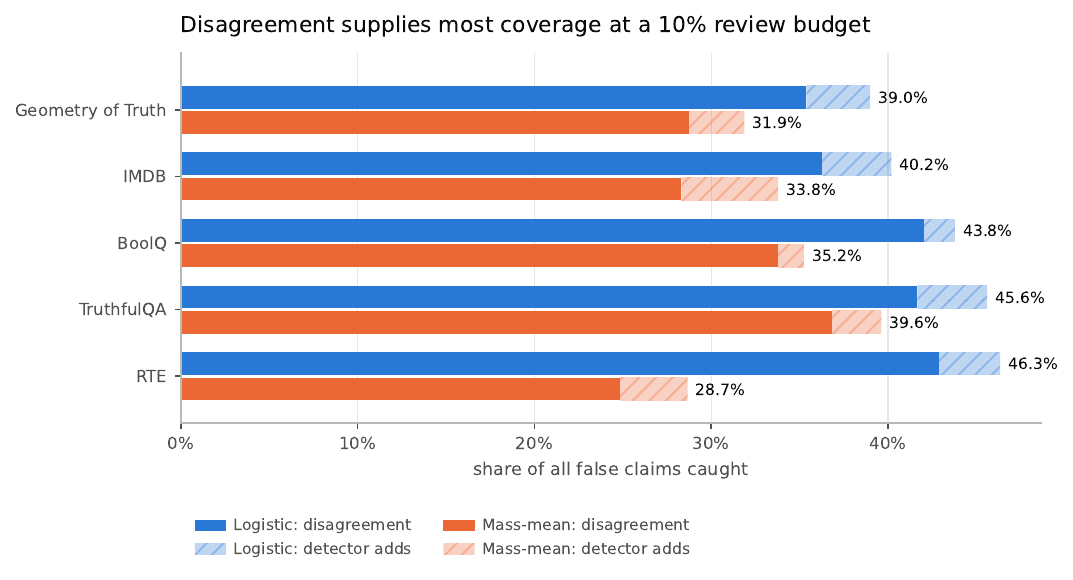}
\caption{Share of all false claims caught by dataset and probe. The solid
segment is exposed by disagreement between the two probes. The hatched segment
is added by reviewing the detector's top-ranked $10\%$ of claims on which both
probes say true. Cells with fewer than $20$ false agreements in the evaluated
split are excluded.}
\label{fig:deployment-coverage}
\end{figure}

Figure~\ref{fig:deployment-coverage} decomposes the false claims caught by a
two-probe review process. The solid segment is exposed by disagreement between
the probes; the hatched segment is added by reviewing the detector's top-ranked
$10\%$ of claims on which both probes say true.

%% file: sections/appendix_pairs.tex
\section{Which Two Models to Pair?}
\label{app:pairs}

Table~\ref{tab:best-pairs} reports the mean and range over pairs in each
dataset; the full-precision set of $498$ pair--dataset cells is included as
\texttt{all\_pairs.csv} in the anonymous supplementary ZIP. Pair-level false agreement spans $0.6\%$ to
$18.4\%$ on Geometry of Truth, $3.2\%$ to $29.0\%$ on RTE, $2.8\%$ to $9.5\%$ on IMDB,
$6.3\%$ to $18.7\%$ on BoolQ, and $1.2\%$ to $24.2\%$ on TruthfulQA.
Detectability varies substantially within datasets too: the in-domain detector
ranges from $0.46$ to $0.84$ across BoolQ pairs and from $0.66$ to $0.90$ on
TruthfulQA; its transferred AUROC ranges from $0.11$ to $0.99$ on Geometry of
Truth. Dataset averages therefore do not predict whether a particular pair's
blind spot is detectable.

The pair with the smallest blind spot need not be the easiest one to audit.
Across datasets, these pairs include at least one highly accurate reader,
consistent with $\FA \leq \min(p_1,p_2)$; on Geometry of Truth the best pair
has only ten false agreements in its test split, too few to score the detector.
Selecting for low false agreement and selecting for a detectable residual are
distinct objectives.

Selecting the most decorrelated pair can increase false agreement: on Geometry
of Truth it yields $4.9\%$ versus $0.6\%$ for the best pair, and on BoolQ
$12.8\%$ versus $6.3\%$. These pairs are genuinely less correlated, but pair
weaker readers, so the increase in $p_1p_2$ outweighs the reduction in the
correlation term. Cross-family selection fares little better. The best
cross-family pair matches or beats the best same-family pair in four of five
datasets, but shuffling family labels while holding counts fixed reproduces
most of this gap; none of the five datasets rejects the shuffled comparison
($p$-values $0.28$--$0.998$). Neither diversity nor family identity reliably
identifies the best pair.

\input{tables/best_pairs.tex}

Two conventions are worth restating here, because they are easy to misread from
the main table alone. First, $p_1$ and $p_2$ are the two probes' error rates on
false claims, meaning the share of false claims each reads as true, and FA is
the joint rate, the share of false claims that \emph{both} read as true. FA is a
probability, not a ratio of the two: it is bounded above by $\min(p_1, p_2)$ and
below by $\max(0,\, p_1 + p_2 - 1)$, and the overlap expressed as a fraction of
the two probes' combined errors is a separate quantity reported in
Section~\ref{sec:discussion}.

Second, the pair is unordered. False agreement, $\rho$ and the router's coverage
are all symmetric in the two models, so there is no overseer and no overseen
model in any of these numbers; the grid was run in both directions and
deduplicated. We therefore list the model with the higher error rate first,
and a reader should not read the first position as a judge or the second as the
model being judged.

The CSV contains the dataset, unordered model pair (listed higher-error first),
family relation, individual false-positive rates, false-agreement rate, error
correlation, and leave-one-dataset-out detector AUROC with its test positive
count.

%% file: tables/best_pairs.tex
\begin{table}[!htbp]
\centering
\small
\setlength{\tabcolsep}{3pt}
\caption{False agreement and its detectability across model pairs, mass-mean probes. Each entry is the mean over pairs with the range in brackets. FA is the share of false claims both probes accept; $\rho$ is the correlation between the two probes' errors. The remaining columns give AUROC for ranking false agreements among agreed claims: the label-free margin from the two outputs (thresholded to the probes' positive rate) and from the two probes; the two models' logistic truth probes, summed, and the supervised false-agreement detector, both fitted within the dataset; and the detector fitted on the other four. Cells need at least 20 false agreements to score; RTE has none in-domain and is omitted. Table~\ref{tab:unsup} gives logistic probes; Appendix~\ref{app:pairs} lists every pair. Per row, the best of the four in-domain detection means is in bold and the second underlined. The transferred detector (italic) is scored on held-out cells with at least 20 false agreements, a different set, so it is not ranked against the others.}
\label{tab:best-pairs}
\resizebox{\textwidth}{!}{%
\begin{tabular}{lrcccccc@{\hspace{14pt}}c}
\toprule
 & & & & \multicolumn{2}{c}{Margin, no labels} & \multicolumn{2}{c}{Target labels} & \multicolumn{1}{c}{Source labels} \\
\cmidrule(lr){5-6}\cmidrule(lr){7-8}\cmidrule(l){9-9}
Dataset & Pairs & FA (\%) & $\rho$ & Outputs & Probes & Truth probe & Detector & \textit{Transferred} \\
\midrule
GoT        & 105 & 4.0 [0.6--18.4] & +0.36 & 0.64 [0.37--0.91] & 0.66 [0.52--0.85] & \textbf{0.98} [0.93--1.00] & \underline{0.90} [0.76--0.99] & \textit{0.68 [0.11--0.99]} \\
IMDB       & 78 & 6.1 [2.8--9.5] & +0.43 & 0.69 [0.47--0.89] & 0.79 [0.65--0.89] & \textbf{0.89} [0.85--0.94] & \underline{0.87} [0.76--0.94] & \textit{0.82 [0.41--0.96]} \\
BoolQ      & 105 & 11.4 [6.3--18.7] & +0.23 & 0.64 [0.51--0.76] & 0.69 [0.55--0.77] & \textbf{0.81} [0.70--0.89] & \underline{0.76} [0.46--0.84] & \textit{0.58 [0.47--0.80]} \\
TruthfulQA & 105 & 10.2 [1.2--24.2] & +0.19 & 0.53 [0.44--0.69] & 0.69 [0.53--0.85] & \textbf{0.93} [0.87--0.98] & \underline{0.80} [0.66--0.90] & \textit{0.54 [0.37--0.74]} \\
\bottomrule
\end{tabular}}
\end{table}